\documentclass{article} 
\usepackage{nips11submit_e,times}
\pdfoutput=1

\usepackage{graphicx}  
\usepackage{subfigure}  
\usepackage{amsmath}  
\usepackage{amssymb}  
\usepackage{algorithm}
\usepackage{algorithmic}
\usepackage{wrapfig}

\newcommand{\Real}{\mathbb{R}}

\newcommand{\ie}{i.e. }
\DeclareMathOperator*{\argmin}{arg\,min}

\title{MIS-Boost: Multiple Instance  Selection Boosting}

\author{
Emre Akbas ~~~~~~~~~ Bernard Ghanem ~~~~~~~~~ Narendra Ahuja
\\
Department of Electrical and Computer Engineering\\
Computer Vision and Robotics Lab \\
Beckman Institute for Advanced Science and Technology \\
University of Illinois at Urbana-Champaign, IL USA 61801\\
\texttt{\{eakbas,bghanem2,ahuja\}@vision.ai.uiuc.edu} \\
}

\nipsfinalcopy 

\begin{document}

\maketitle

\begin{abstract}
In this paper, we present a new multiple instance learning (MIL) method, called MIS-Boost, which learns
discriminative instance prototypes by explicit instance selection in a
boosting framework. Unlike previous instance selection based MIL
methods, we do not restrict the prototypes to a discrete set of training instances but allow them to take arbitrary values in the instance feature space. We also do not restrict the total number of prototypes and the number of selected-instances per bag; these quantities are completely data-driven. We show that MIS-Boost outperforms state-of-the-art MIL methods on a number of benchmark
datasets. We also apply MIS-Boost to large-scale image classification,
where we show that the automatically selected prototypes map to visually meaningful image regions.


\end{abstract}

\section{Intoduction} \label{sec:introduction}

Traditionally, supervised learning algorithms require labeled training data,
where each training instance is given a specific class label. The performance
and learning capability of such algorithms are impacted by the correctness of
these instance labels especially when they are obtained through human
interaction. In some applications (e.g. object detection, semantic segmentation,
and activity recognition), ambiguities in human labeling may arise (e.g. when
detection bounding boxes are not accurately sized or positioned). Traditional
supervised learning methods cannot easily resolve these label ambiguities, which
are inherently handled by multiple instance learning methods. These latter
methods are based on a significantly weaker assumption about the underlying
labels of the training instances. They do not assume the correctness of each
individual instance label, yet they assume label correctness at the level of
groupings of training instances. Multiple instance learning (MIL) can be viewed
as a weakly supervised learning problem where the labels of sets of instances
(known as \emph{bags}) are given, while the labels of the instances in each bag
are unknown. In a typical binary MIL setting, a negative bag contains instances
that are all labeled negative, while a positive bag contains at least one
instance labeled positive. Since the instance labels in positive bags are
unknown, a MIL classifier seeks an optimal labeling scheme for the training
instances so that the resulting labels of the training bags are correct.

Due to their ability to handle incomplete knowledge about instance labels of
training data, MIL methods have gained significant attention in the machine
learning and computer vision communities. In fact, the MIL learning framework
manifests itself in numerous applications that span from text categorization
\cite{Andrews2003} and drug activity recognition \cite{Dietterich1997} to many
vision applications, where an image (or video) is represented as a bag of
instances (interest points, patches, or image segments), only a subset of which
are meaningful for the task in question (e.g. image classification). Recently,
it has been shown that MIL methods can achieve state-of-the-art performance in a
multiplicity of vision applications including content-based image retrieval
\cite{Zhang2002,Cholleti2006,Vijayanarasimhan2008}, image classification
\cite{Fu2010a,Chen2006,Foulds2008}, activity recognition \cite{Stikic2009},
object tracking \cite{Babenko2010,Leistner2010}, object detection
\cite{milboost:nips2007,Zhang2008}, and image segmentation
\cite{Vezhnevets2010}. In fact, label ambiguity and incompleteness lead to
significant challenges in the MIL framework, especially when positive bags are
dominated by negative instances (e.g. an image of an airplane dominated by
patches of sky). It is not surprising to see empirical evidence that the use of
traditional supervised learning methods in MIL problems often leads to reduced
accuracy \cite{Ray2005}. Consequently, much effort has been done to develop
effective learning methods that exploit the structure of MIL problems. We will
give a brief overview of the most recent and popular MIL methods next.

\subsection*{Related Work} \label{sec:introduction}
During the past two decades, many MIL methods have been proposed with a
significant interest in MIL emerging in recent years especially within the
machine learning and vision communities. The work in \cite{Keeler1990} is one of
the earliest papers that address the MIL problem, whereby it was cast in the
framework of recognizing hand-written numerals. Over the next twenty years, MIL
literature has abound with algorithms that differ in two main respects:
\textbf{(1)} the level at which the labeling is determined: instance-level
(bottom-up) or bag-level (top-down) and \textbf{(2)} the type of data modeling
assumed, i.e. generative vs. discriminative. 

\vspace{-2mm}
\paragraph{\textbf{(1). Bottom-up vs. Top-down}:} While most MIL approaches address the problem of predicting the class of a bag directly without inferring the labels of the instances that belong to this bag, some approaches use max margin techniques to do this inference \cite{Andrews2003,Cheung2006,Zhou2007a}. The latter approaches exploit the fact that the instances of negative bags have negative labels ($-1$) and at least one instance in each positive bag has a positive label ($+1$). For example, a MIL version of SVM is proposed in \cite{Andrews2003}, where the traditional SVM optimization problem is transformed into a mixed-integer program and subsequently solved by alternating between solving a traditional SVM problem and heuristically choosing the positive instances for each positive bag.

\vspace{-2mm}
\paragraph{\textbf{(2). Generative vs. Discriminative}:} Some MIL approaches are
generative in nature, since they assume that the underlying instances conform to
a certain structure. An early generative approach is based on finding an optimal
hyper-rectangle discriminant in the instance space \cite{Dietterich1997}. Other
prominent generative MIL approaches are based on the notion of diverse density
(DD) \cite{Maron1998}. These approaches seek a ``concept" instance\footnote{This
instance does not have to be one of the instances in the training set.} that is
\emph{close} to at least one instance of each positive bag and \emph{far} away
from all the instances in the negative bags. In other words, a concept instance
is a vector in instance space that best describes the positive bags and
discriminates them from the negative bags. The existence of such a concept
assumes that positive instances are compactly clustered and well separated from
negative instances. Such an assumption is strict and does not always
hold in natural data, which tends to be multi-modal. DD-based MIL approaches
compute this optimal concept by formulating the problem in a maximum likelihood
framework using a noisy-OR model of the likelihood. Improvements on the original
DD formulation have been made, where the EM algorithm is used to find the
concept instance in \cite{Zhang2002} and multiple concepts are estimated in
\cite{Rahmani2008}. Moreover, standard supervised learning techniques, such as
kNN, linear and kernel SVM, AdaBoost, and Random Forests, have been adapted to
the MIL problem, thus, leading to citation kNN \cite{Wang2000}, MI-kernel
\cite{Gartner2002}, MIGraph/miGraph \cite{Fu2010a}, mi/MI/DD-SVM
\cite{Andrews2003,Chen2004}, MI-Boost \cite{milboost}, MI-Winnow
\cite{Cholleti2006}, MI logistic regression \cite{Fu2008}, and most recently
MIForests \cite{Leistner2010}. Furthermore, some MIL approaches actively seek
instances in the training set (denoted prototypes) that carry discriminative
power between the positive and negative classes. In what follows, we will denote
these as instance selection MIL methods. Such approaches transform the original
feature space into another space defined by the selected prototypes (e.g. using
bag-to-instance similarities) and subsequently apply standard supervised
learning techniques in the new space. In \cite{Chen2006}, all training instances
are selected to be prototypes, while only one instance per bag is systematically
initialized, greedily updated, and selected in \cite{Fu2009,Fu2010a}.

Our proposed instance selection method (dubbed Multiple Instance Selection Boost
or MIS-Boost) is inspired by the instance selection MIL methods mentioned above
(e.g. MILES \cite{Chen2006,Foulds2008} and MILIS \cite{Fu2010a}) and the
MI-Boost algorithm in \cite{milboost}. It was hinted earlier that instance
selection MIL methods comprise two fundamental stages. \textbf{(i)} In the
representation stage, the original training bags are represented in a new feature space determined by the selected prototypes. \textbf{(ii)} In the classification stage, a supervised learning technique is used to build a classifier in the new feature space to optimize a given classification cost. Most of these methods treat the two stages independently and sequently, in such a way that representation (i.e. prototype selection) is unaffected by class label distribution. Here, MILIS is an exception, since it iteratively selects prototypes \emph{from the training set} to minimize classification cost. This selection is further restricted, since only one prototype is selected per training bag. We consider this to be a restriction because we believe that the prototype selection process should be data dependent. For example, in the case of image classification, some ``simple" object classes (e.g. airplane) may yield a smaller number of prototypes than other more ``complex" classes (e.g. bicycle). 

As compared to previous methods, prototypes selected by MIS-Boost do not
necessarily belong to the given training set and the number of these prototypes
is not predefined, since they are determined in a data-driven fashion
(boosting). Since the search space for prototype instances is no longer limited,
more discriminative and possibly fewer prototypes can be \emph{learned}. This
learning process directly involves minimization of the final classification
cost. As such, MIS-Boost \emph{learns} a new representation based on the
estimated prototypes, in a boosting framework. This leads to an iterative
algorithm, which learns prototype-based base classifiers that are linearly
combined. At each iteration of MIS-Boost, a prototype is learned so that it
maximally discriminates between the positive and negative bags, which are
themselves weighted according to how well they were discriminated in earlier
iterations. The number of prototypes is determined in a data-driven way by
cross-validation. Experiments on benchmark datasets show that MIS-Boost achieves state-of-the-art performance. When
applied to image classification, MIS-Boost selects prototypes that map to
meaningful image segments (e.g. class specific object parts).

In Section \ref{sec:algorithm}, we give a detailed description of the MIS-Boost algorithm including our proposed instance selection/learning method. We show that MIS-Boost achieves or improves upon state-of-the-art results on benchmark MIL datasets and popular image classification datasets in Section \ref{sec:experiments}.

\section{Proposed Algorithm} \label{sec:algorithm}

Given a training set $T = \{(B_1,y_1), (B_2,y_2), \dots, (B_N, y_N)\}$ where
$y_i \in \{-1, +1\}\;\;\allowbreak \forall i$,
$B_i$ represents the $i^{\text{th}}$ bag and $y_i$ its label, our goal is to learn a
bag-classifier $F: B \rightarrow \{-1, +1\}$. Each bag consists of an arbitrary
number of instances. The number of instances in the $i^{\text{th}}$ bag is denoted by
$n_i$, so we have $B_i = \{ \vec{\mathbf{x}}_{i1}, \vec{\mathbf{x}}_{i2}, \dots, \vec{\mathbf{x}}_{in_i}\}$, where each instance
$\vec{\mathbf{x}}_{ij} \in \Real^N\;\;\forall
i,j$. We propose the following additive model as our bag classification function:

\begin{equation}
F(B) = \mathrm{sign}\left(\sum_{m=1}^M f_m (B)\right),
\end{equation}

\noindent where each $f_m(B)$, called  a base classifier, is associated with a prototype instance $\vec{\mathbf{p}}_m \in \Real^N$. The function $f_m : B \rightarrow [-1,1]$
is a bag classifier, like $F$, and it returns a score between $-1$ and
$1$, which quantifies the ``existence" of the prototype instance $\vec{\mathbf{p}}_m$ within bag $B$. The ``existence" of $\vec{\mathbf{p}}_m$ within bag $B_i$ is determined by the distance from
$\vec{\mathbf{p}}_m$ to the closest region to $\vec{\mathbf{p}}_m$ within $B_i$, that is:

\begin{equation}
\label{eq:instance_to_bag_dist}
D(\vec{\mathbf{p}}_m, B) = \min_{j} ~d(\vec{\mathbf{p}}_m, \vec{\mathbf{x}}_{ij}),
\end{equation}

\noindent where $d(\cdot, \cdot)$ is a distance function between two instances,
which we take to be Euclidean.  We denote $D(\cdot, \cdot)$ as the
instance-to-bag distance function. Here, we note that this instance-to-bag
distance is used in other instance selection MIL methods (e.g. MILES and MILIS);
however, the prototype $\vec{\mathbf{p}}_m$ in these methods is restricted to a
discrete subset of the training samples. By removing this restriction on
$\vec{\mathbf{p}}_m$ and allowing it to take arbitrary values in $\Real^N$, more
discriminative and possibly fewer prototypes need to be \emph{learned}. The
function $f_m(\cdot)$ computes the instance-to-bag distances first, and classifies
the bags using these distances. Although $f_m(\cdot)$ can take any suitable
form, we opt to use the simple scaled and shifted sigmoid function,
parameterized by $\beta_0$, $\beta_1$, and $\vec{\mathbf{p}}_m$.

\begin{equation}
f_m(B) = \frac{2}{1+e^{-(\beta_1 D(\vec{\mathbf{p}}_m, B) + \beta_0)}}-1,
\end{equation}


Since $F$ is an additive model, we use additive boosting to learn its base classifiers
\cite{statistical_view_of_boosting}. We call our algorithm multiple instance
selection boosting, or MIS-Boost for short. One of the main reasons for choosing
boosting for classification is its ability to select a suitable number of
prototypes, in a data-driven fashion. By performing cross-validation, not only
does the classifier avoid overfitting on the training set, but it also
automatically determines the number of base classifiers (i.e. the number of
prototypes) needed to form $F$. Note that other instance selection methods
predefine or fix the number of prototypes that are used. Among the many variants
of boosting, we choose Gentle-AdaBoost for its numerical stability properties
\cite{statistical_view_of_boosting}.


\subsection{Learning base classifier $f_m$}
\label{sec:learning_fm}
At each iteration of Gentle-AdaBoost, a weighted least-squares problem must be solved (step 2(a), Algorithm 4 in \cite{statistical_view_of_boosting}).  In our formulation, the following error should be minimized:

\begin{equation}
\label{eq:weak_classifier_cost}
\argmin_{\vec{\mathbf{p}}_m, \beta_0, \beta_1} \varepsilon_m\;\;\;\text{where}\;\;
\varepsilon_m  = \sum_{i=1}^N w_i \left(y_i - \frac{2}{1+e^{-(\beta_1 D(\vec{\mathbf{p}}_m, B_i) +
\beta_0)}}+1 \right)^2
\end{equation}

\noindent Here, $w_i$ is the weight of the $i^{\text{th}}$ bag at the current
iteration. The main difficulty in optimizing the cost function above is the fact
that the instance-to-bag distance term $D(\vec{\mathbf{p}}_m,B)$ involves the
non-differentiable ``min" function. It is this same function that forces other
instance selection methods (e.g. MILES and MILIS) to restrict the prototype
search space to a subset of the training samples. For example, MILES considers
all training samples as valid prototypes, thus, making learning the classifier
($\ell_1$ SVM) significantly computationally expensive. On the other hand, MILIS
takes a brute-force approach to prototype selection by greedily choosing one
instance from each training bag as a valid prototype. Although selection is done
so that an overall classification cost is iteratively reduced, this selection
strategy highly restricts the feasible prototype space. To alleviate the problem
of non-differentiability in our formulation, we replace ``min" in
$D(\vec{\mathbf{p}}_m,B_i)$ with a differentiable approximation (known as
``soft-min") to form the soft-instance-to-bag distance
$\tilde{D}(\vec{\mathbf{p}}_m, B_i)$. By setting $\alpha$ to a large positive
constant, we have:


\begin{equation}
D(\vec{\mathbf{p}}_m, B_i) \approx \tilde{D}(\vec{\mathbf{p}}_m, B_i) = \sum_{j=1}^{n_i} \pi_j d(\vec{\mathbf{p}}_m, \vec{\mathbf{x}}_{ij}), \;\;\;\text{where}\;\; \pi_j = \frac{\displaystyle e^{-\alpha d(\vec{\mathbf{p}}_m, \vec{\mathbf{x}}_{ij})}}{\displaystyle
\sum_{k=1}^{n_i} e^{-\alpha d(\vec{\mathbf{p}}_m, \vec{\mathbf{x}}_{ik})}}.
\end{equation}

%



Replacing $D(\vec{\mathbf{p}}_m,B)$ with $\tilde{D}(\vec{\mathbf{p}}_m,B)$ in
Eq. \eqref{eq:weak_classifier_cost} renders the cost function differentiable,
allowing for gradient descent optimization. However, it is not a convex cost
function, so there is a risk of settling into undesirable local minima. To alleviate this problem, we allow for multiple initializations of $\vec{\mathbf{p}}_m$. Preferably, these initialization points should be sampled from
the entire instance feature space. For this purpose, we cluster all the
training instances using $k$-means, and use the cluster centers as
initialization points for $\vec{\mathbf{p}}_m$. We minimize the cost in Eq. \eqref{eq:weak_classifier_cost} using
coordinate-descent. We start by initializing $\vec{\mathbf{p}}_m$ to a cluster center and
optimize over the $(\beta_0,\beta_1)$ parameters. Then, we fix these parameters and optimize over $\vec{\mathbf{p}}_m$. We iterate this procedure until convergence; that is,
the difference between successive errors becomes smaller than a given threshold.
The overall algorithm used to learn a base classifier is summarized in Algorithm
\ref{algo:learning_weak_classifier}.

\begin{algorithm}[htb]
\caption{Learning $f_m$ (Pseudo-code for learning a base classifier)}
\label{algo:learning_weak_classifier}
\begin{algorithmic}
\REQUIRE  Training set $\{(B_1,y_1), (B_2,y_2), \dots, (B_N, y_N)\}$, Weights
$w_i, i=1,2,\dots,N$, cluster centers $\{c_1, c_2, \dots, c_K\}$, Error tolerance Tol.
\ENSURE Base classifier $f_m(x)$
\STATE // Initialize $\vec{\mathbf{p}}_m$ to each cluster center
\FOR{$\vec{\mathbf{p}}_m^0= c_1, c_2, \dots, c_K$}
\STATE error(-1) $\gets \infty$
\STATE $(\beta_0^0, \beta_1^0) \gets \argmin \tilde{\varepsilon}_m |_{(\vec{\mathbf{p}}_m =
\vec{\mathbf{p}}_m^0)}$ \COMMENT{Fix $\vec{\mathbf{p}}_m$ and minimize over $\beta$'s}
\STATE error(0) $\gets \tilde{\varepsilon}_m (\vec{\mathbf{p}}_m^0, \beta_0^0, \beta_1^0)$
\STATE $t \gets 0$
\WHILE{$|$error($t+1$) $-$ error($t$)$|\ge$ Tol}
\STATE $t \gets t + 1$
\STATE $\vec{\mathbf{p}}_m^t \gets \argmin_{\vec{\mathbf{p}}_m} \tilde{\varepsilon}_m|_{(\beta_0 =
\beta_0^{t-1}, \beta_1 = \beta_1^{t-1})}$ \COMMENT{Fix $(\beta_0,\beta_1)$ and minimize
over $\vec{\mathbf{p}}_m$}
\STATE $(\beta_0^t, \beta_1^t) \gets \argmin \tilde{\varepsilon}_m |_{(\vec{\mathbf{p}}_m =
\vec{\mathbf{p}}_m^{t-1})}$
\STATE error($t$) $\gets \tilde{\varepsilon}_m (\vec{\mathbf{p}}_m^t, \beta_0^t, \beta_1^t)$
\ENDWHILE
\STATE Keep $(\vec{\mathbf{p}}_m^*, \beta_0^*, \beta_1^*)$ with the least error so far
\ENDFOR
\STATE Set $(\vec{\mathbf{p}}_m, \beta_0, \beta_1) \gets (\vec{\mathbf{p}}_m^*, \beta_0^*, \beta_1^*)$, and  output $f_m$
\end{algorithmic}
\end{algorithm}

\subsection{Determining the number of base classifiers}
As the number of base classifiers increases, Gentle-AdaBoost tends to overfit on the training data. In order to prevent this and determine the number of base classifiers automatically, we perform $4$-fold cross validation, whereby we randomly split the training data into $4$ equal-size pieces and use $3$ pieces for training and the rest for validation. We run the algorithm for a large number of base classifiers and pick the number which gives the least classification error on the validation set. We give the pseudo-code of MIS-Boost in Algorithm \ref{algo:misboost}.


\begin{algorithm}[htb]
\caption{MIS-Boost (Pseudo-code for the MIS-Boost algorithm)}
\label{algo:misboost}
\begin{algorithmic}
\REQUIRE  Training set $\{(B_1,y_1), (B_2,y_2), \dots, (B_N, y_N)\}$, maximum number of
base classifiers $M$, number of clusters $K$
\ENSURE Classifier $F(x)$
\STATE Cluster all instances, $x_{ij}, \;i=1,2,\dots,N; j = n_i$, into $K$
clusters.
\STATE Cluster centers are $\{c_1, c_2, \dots, c_K\}$.
\STATE Split the training set into train-set and validation-set.
\STATE Weights $w_i \gets 1/N$ for $i=1,2,\dots, N$, and $F(x) = 0$.
\FOR{$m=1,2,\dots,M$}
\STATE Learn a base classifier $f_m$ using the algorithm given in
Algorithm
\ref{algo:learning_weak_classifier}.
\STATE Update $F(x) \gets F(x) + f_m(x)$,
\STATE Update $w_i \gets w_i e^{-y_i f_m(B_i)}$ and normalize weights so that
$\sum w_i = 1$.
\STATE Evaluate $F(x)$ on the validation-set, compute validation-error$(m)$.
\ENDFOR
\STATE $M \gets \argmin_m$ (validation-error)
\STATE Output $F(x) = \mathrm{sign} \left( \sum_{i=1}^M f_m(x) \right)$
\end{algorithmic}
\end{algorithm}

\section{Experiments} \label{sec:experiments}
In this section, we evaluate the performance of MIS-Boost on five different
MIL benchmark datasets and two COREL image
classification datasets. We compare our performance to those of the most recent
and state-of-the-art MIL methods available for each dataset. In
another experiment, we use MIS-Boost in a large-scale image classification task
and visualize samples of the instances that are closest to the
learned prototype(s). The results of this experiment suggest that the learned
prototypes are not only discriminative but also visually meaningful, that is
they are similar to the parts of image that are relevant for classification.

\subsection{Benchmark MIL datasets}
The drug activity prediction datasets, ``Musk1" and ``Musk2" described in
\cite{Dietterich1997},  and the image datasets, ``Elephant", ``Fox", ``Tiger"
introduced in \cite{Andrews2003} have been widely used and have become standard benchmark datasets
for MIL methods. For each dataset, we perform
$10$-fold cross validation and report the average per-fold test classification
accuracy. This is the standard way of reporting results on these datasets.
In all our experiments in this section and the two following sections, we set
the number of clusters $K=100$, and the maximum number of base learners, or
prototypes, to $M=100$.

We report our results in Table \ref{tab:mil_benchmark_results}, where we list the results of the most recent and state-of-the-art MIL methods. To the
best of our knowledge, this table gives the most comprehensive comparison
between MIL methods on the benchmark datasets.
Clearly, MIS-Boost outperforms
other methods in all datasets except the ``Tiger" class. There, we have the second best
accuracy with only a $0.5\%$ difference with the top performing method, miGraph
\cite{Zhou2009}.  Among all the methods in Table
\ref{tab:mil_benchmark_results}, MIS-Boost is the most similar to MILES and
MILIS, as they are also instance-selection-based methods. Except on ``Musk1",
our algorithm significantly outperforms these two methods. We believe that this
improvement is largely due to the fact that our method, in contrast to MILES and
MILIS, does \emph{not} restrict the prototypes to a subset of the training
instances, as we discussed in Section \ref{sec:learning_fm}.

\begin{table}
\caption{
\label{tab:mil_benchmark_results}
Percent classification accuracies of MIL algorithms on benchmark MIL datasets.
Best results are marked in bold fonts.
}
\centering
\begin{tabular}{|l|c|c|c|c|c|}
\hline
Method & Musk1 & Musk2 & Elephant & Fox & Tiger \\
\hline
\hline
MIS-Boost &  \bf 90.3 & \bf 94.4 & \bf 89.0 & \bf 80.0  & 85.5 \\
\hline
MIForest\cite{Leistner2010} & 85 & 82 & 84 & 64 & 82 \\
MIGraph\cite{Zhou2009} & 90.0 & 90.0 & 85.1 & 61.2 & 81.9 \\
miGraph\cite{Zhou2009} & 88.9 & 90.3 & 86.8 & 61.6 & \bf 86.0 \\
MILBoost\cite{milboost} & 71 & 61 & 73 & 58 & 58\\
EM-DD\cite{Zhang2002} & 84.8 & 84.9 & 78.3 & 56.1 & 72.1 \\
DD\cite{Chen2004} & 88.0 & 84.0 & N/A & N/A & N/A \\
MI-SVM\cite{Andrews2003} & 77.9 & 84.3 & 81.4 & 59.4 & 84.0 \\
mi-SVM\cite{Andrews2003} & 87.4 & 83.6 & 82.0 & 58.2 & 78.9 \\
MILES\cite{Chen2006} & 88 & 83 & 81 & 62 & 80 \\

MILIS\cite{milis:pami2010} & 88 & 83 & 81 & 62 & 80 \\

MI-Kernel\cite{Gartner2002} & 88 & 89 &  84 & 60 & 84 \\
AW-SVM\cite{GehlerChapelle_AISTATS2007} & 86 & 84 & 82 & 64 & 83 \\
AL-SVM\cite{GehlerChapelle_AISTATS2007} & 86 & 83 & 79 & 63 & 78\\
MissSVM\cite{Zhou2007a} & 87.6 & 80.0 & N/A & N/A & N/A\\
\hline
\end{tabular}
\end{table}

\subsection{COREL dataset}
\begin{figure}
\centering
\includegraphics[scale=0.45]{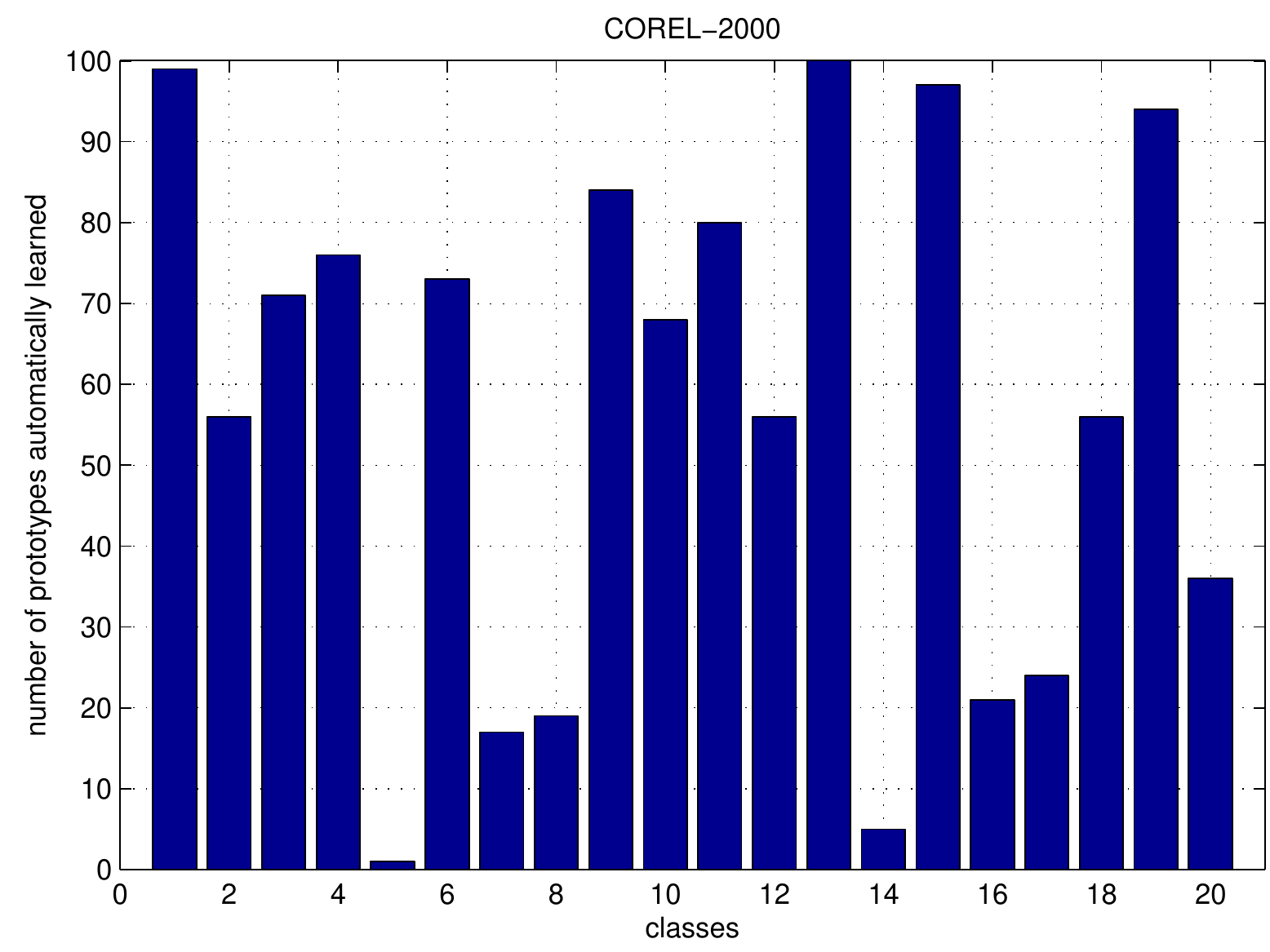}
\caption{Number of base learners, or prototypes, per class as determined by MIS-Boost on
the COREL-2000 dataset.}
\label{fig:opt_num_weaks}
\end{figure}

The COREL-2000 image classification dataset \cite{Chen2006} contains $2000$
images in $20$ classes. COREL-1000 is just a subset of this dataset, which
contains the first $10$ classes. We use the same features and experimental
settings as in \cite{Chen2006}, and train one-vs-all MIS-Boost classifiers to deal with 
the multiclass case.  The results of MIS-Boost and three most recent,
state-of-the-art methods are given in Table \ref{tab:corel}. Our method
outperforms the other methods on both datasets. 
To illustrate the data-driven nature of our algorithm, we give the number of
prototypes learned per class in Figure \ref{fig:opt_num_weaks}. 

\begin{table}
\caption{Percent classification accuracies on the COREL-1000 and COREL-2000
datasets.}
\label{tab:corel}
\centering
\begin{tabular}{|l|c|c|}
\hline
Method & COREL-1000 & COREL-2000 \\
\hline
\hline
MIS-Boost & \bf 84.2 & \bf 70.8 \\
\hline
MILIS\cite{milis:pami2010} & 83.8 & 70.1 \\
MILES\cite{Chen2006} & 82.3 & 68.7 \\
MIForest\cite{Leistner2010} & 82 & 69 \\
\hline
\end{tabular}
\end{table}

\subsection{PASCAL VOC 2007}
The image categorization task of PASCAL VOC is inherently a MIL problem since
the label of an image indicates the existence of at least
one object of that label class within the image. However, to the best of our
knowledge, no MIL results have been reported on this dataset. This is probably
because of the large number of images ($10^4$) it contains. If we assume that each
image has a few hundred instances, then the total number of instances is in
the order of millions. Instance selection based methods like MILES would
easily run into memory problems. In this section, we evaluate the performance of
MIS-Boost on this large-scale image classification dataset, and visualize the
selected instances, i.e. those instances that are closest to the learned
prototypes to see if they overlap
with the object(s) of interest. To this end, we run MIS-Boost on three
selected classes from the dataset, ``aeroplane", ``bicycle", and ``tvmonitor".

To decide what type of instances (or features) to use, we did preliminary experiments with
SIFT keypoint/descriptors \cite{sift} and regions obtained by the segmentation
algorithm \cite{akbas:accv2009}. Regions gave better results ($0.55$
average-precision (AP))\footnote{This is the standard performance measure used in
PASCAL VOC.} than the SIFT descriptors ($0.38$ AP) on the ``aeroplane"
class, so we decided to use regions.

This dataset is not only huge but also unbalanced. The ``aeroplane" class has
$442$ positive vs. $9518$ negative; the ``bicycle" class has $482$ positive vs. $9458$
negative; and ``tvmonitor" has $485$ positive vs. $9429$ negative images (bags). This unbalancedness makes finding discriminative instances in the positive bags, among the clutter from the relatively large number of negative bags, quite challenging.

MIS-Boost yields an average-precision (AP) score of $0.55$ for the ``aeroplane" class.
This score is significantly below the state-of-the-art (e.g. $0.76$ in
\cite{Perronnin2010}) on that dataset. We believe that this discrepancy is
largely due to the fact that MIL methods do not model the context (or the
background), while \cite{Perronnin2010} and other similar
approaches do so. Although the MIL approach seems to be the
most appropriate one for the PASCAL dataset (given how the ground-truth is
formed), these results suggest that the context/background information is highly
discriminative. Another reason for the discrepancy might be the instance
features we use, namely the segmentation might fail to capture the object or its parts.
MIS-Boost yields an AP of $0.28$ on ``bicycle" (compared to $0.65$ in
\cite{Perronnin2010}) and $0.36$ on ``tvmonitor" (compared to $0.52$ in
\cite{Perronnin2010}).

\begin{figure}
\centering
\begin{tabular}{ccc}
\includegraphics[scale=.10]{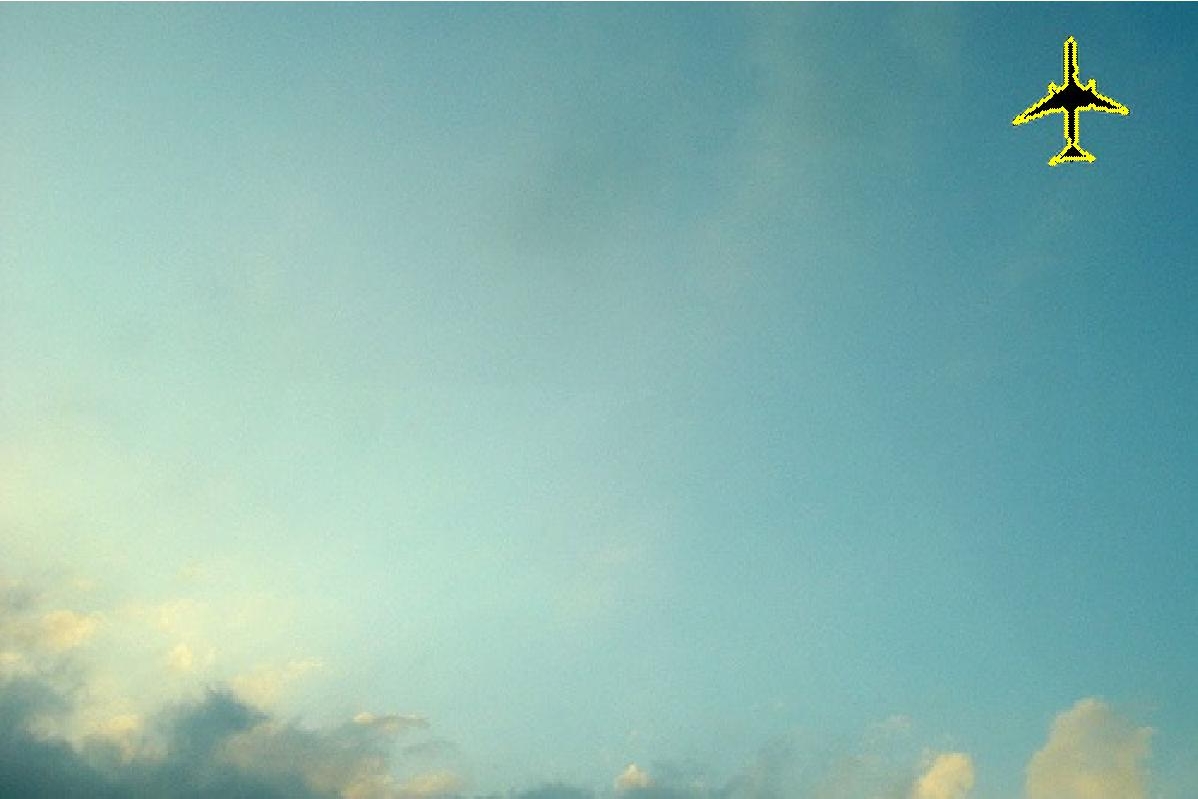}&
\includegraphics[scale=.10]{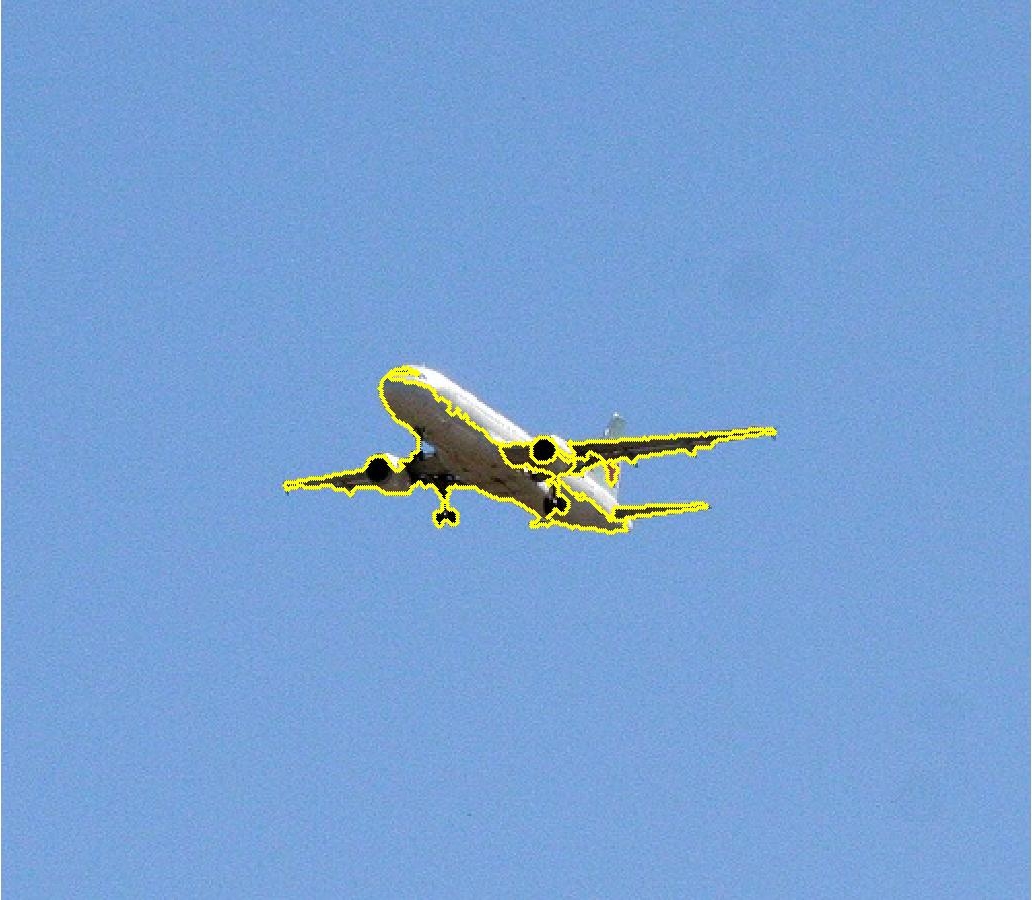}&
\includegraphics[scale=.10]{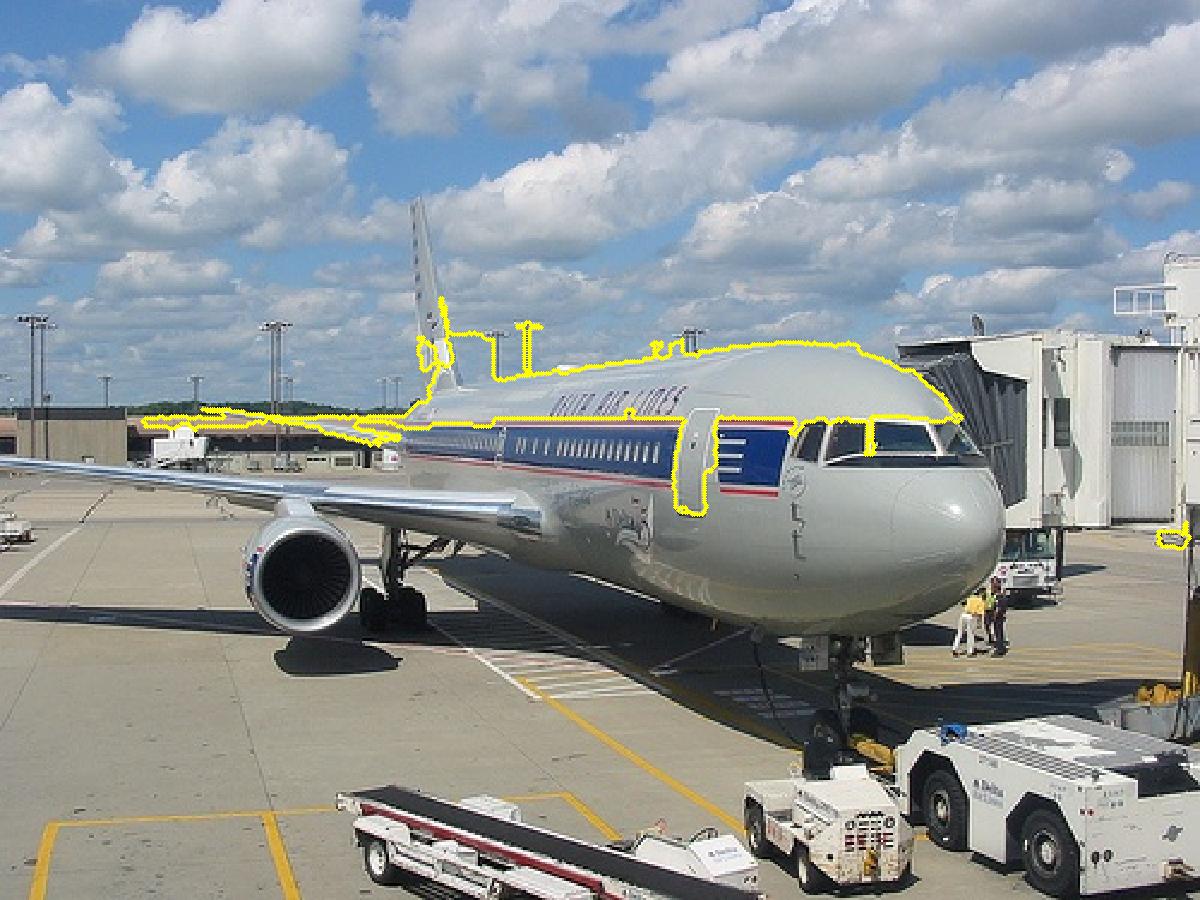}\\
\includegraphics[scale=.10]{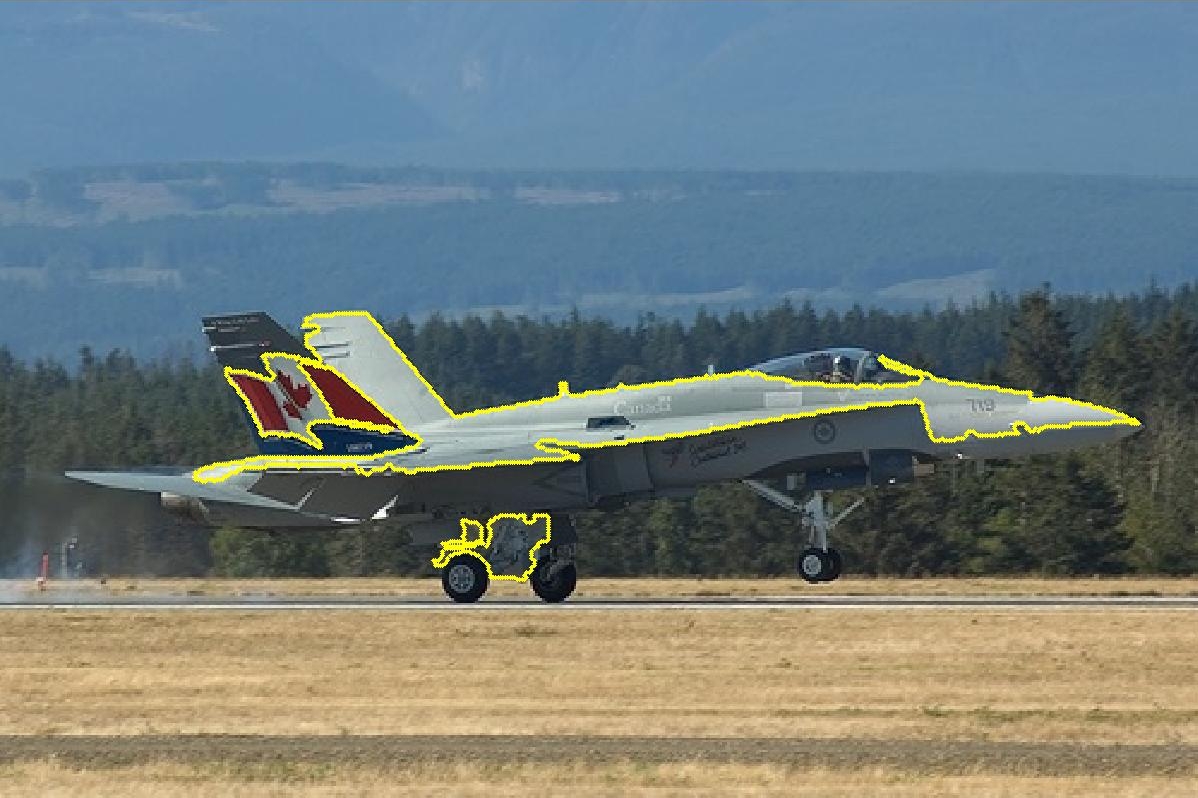}&
\includegraphics[scale=.10]{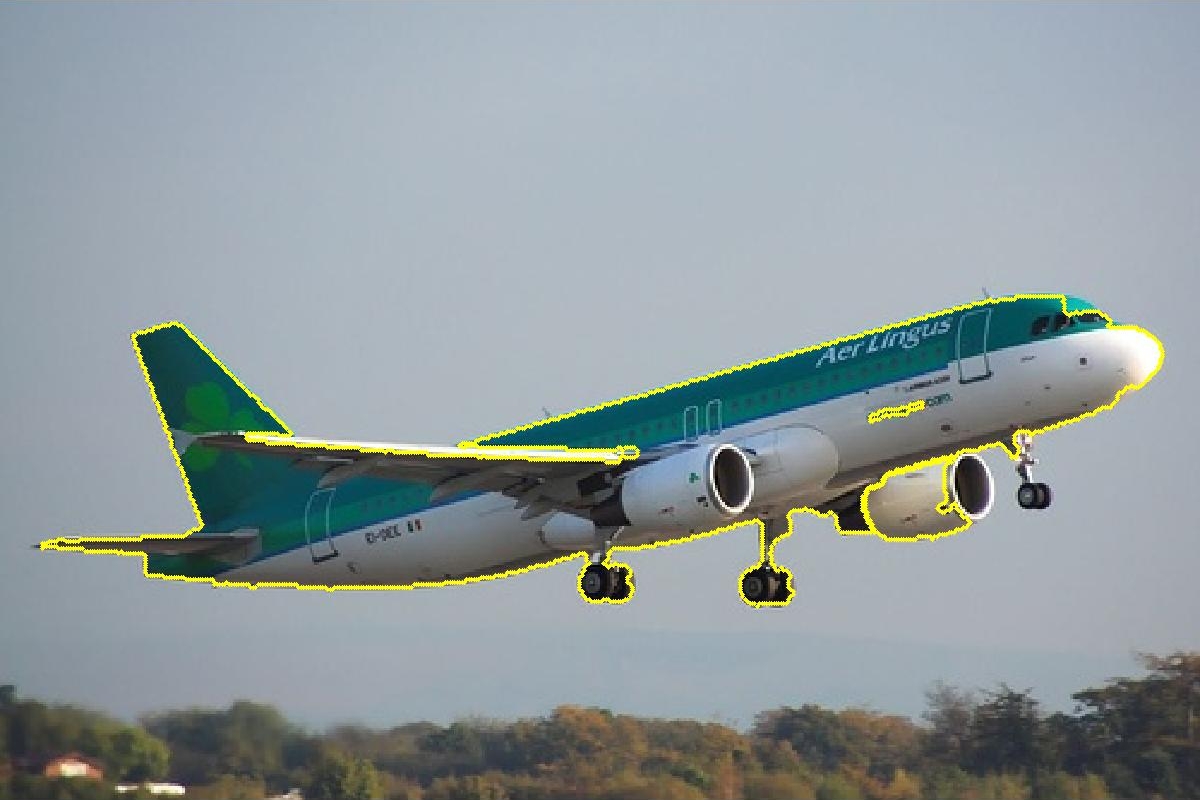}&
\includegraphics[scale=.10]{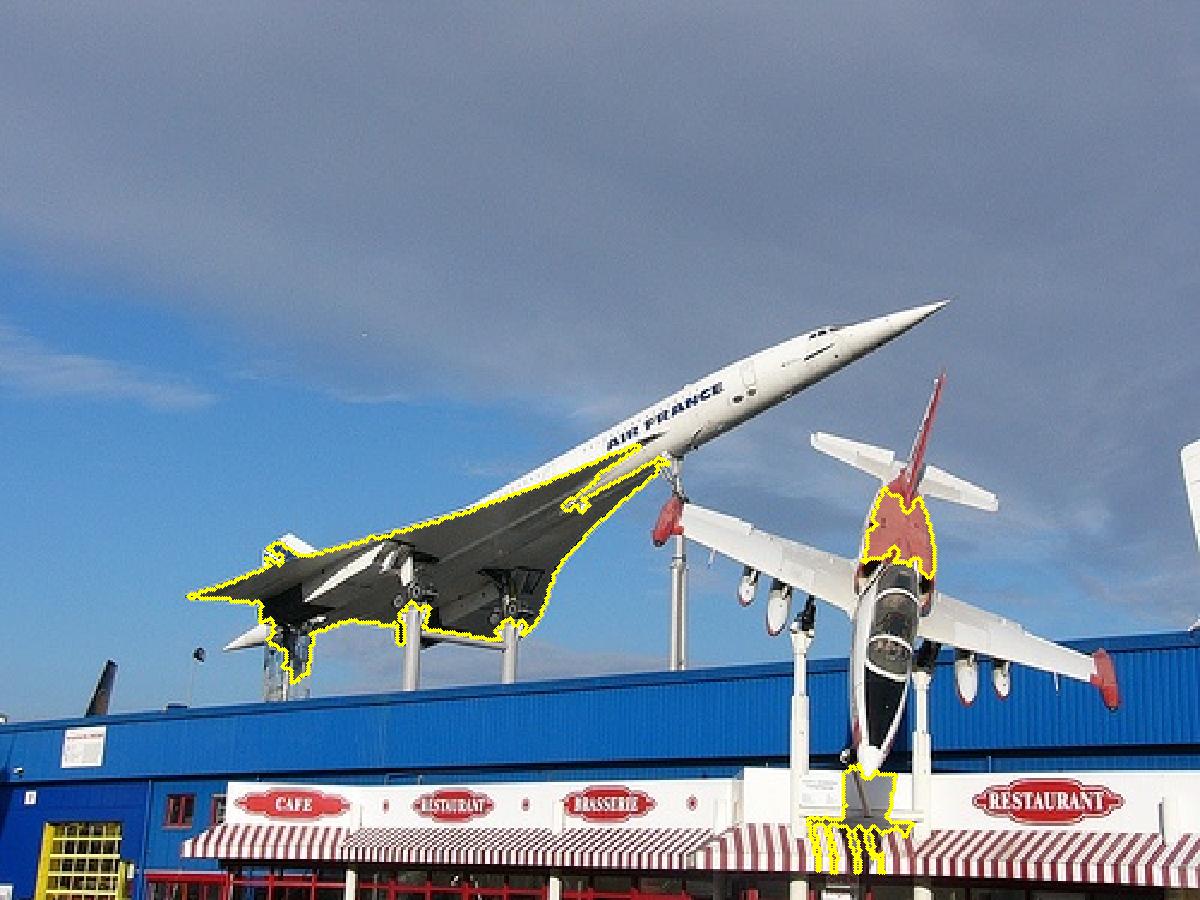}\\
\includegraphics[scale=.10]{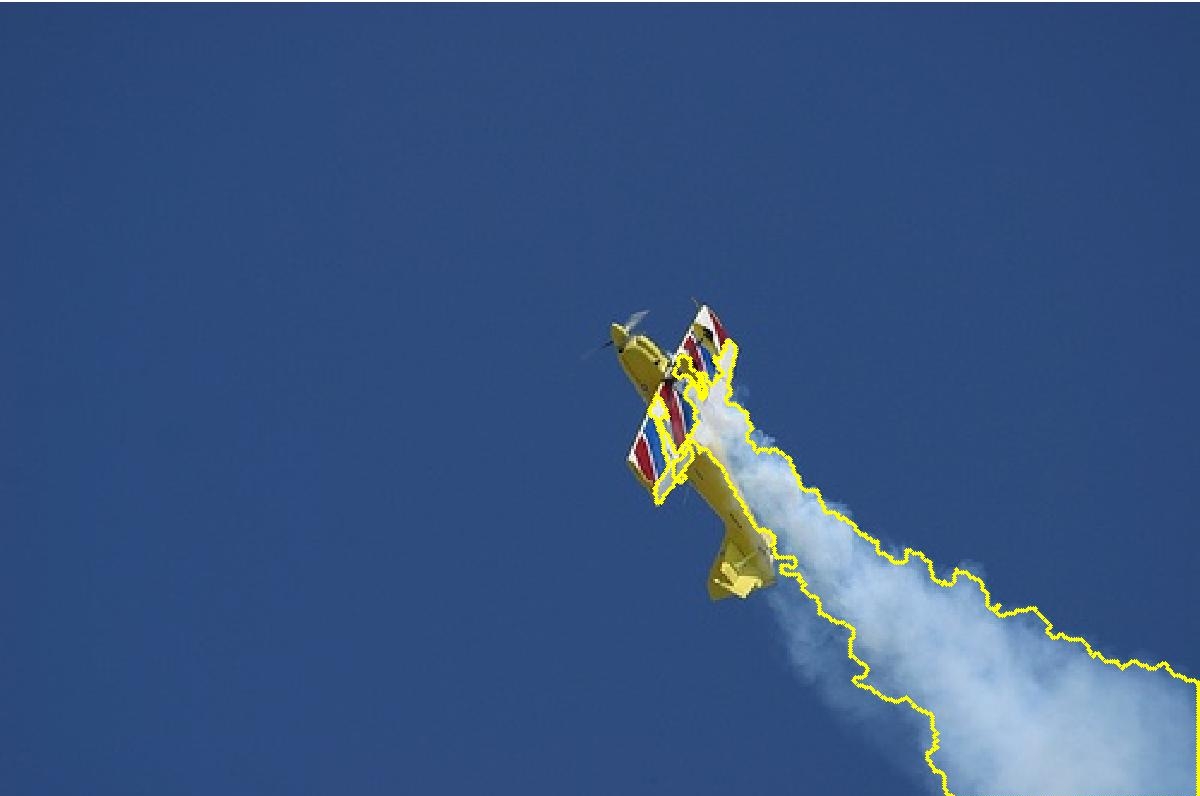}&
\includegraphics[scale=.10]{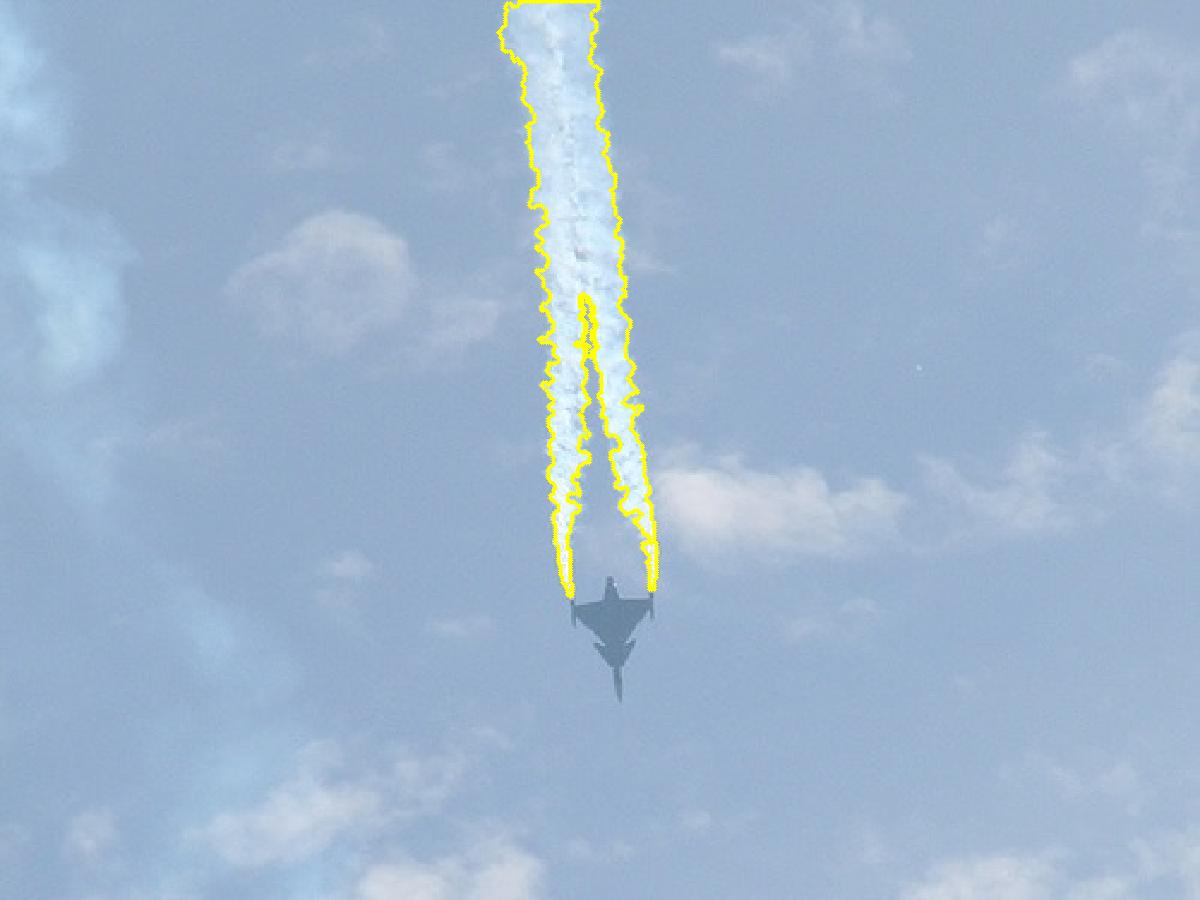}&
\includegraphics[scale=.10]{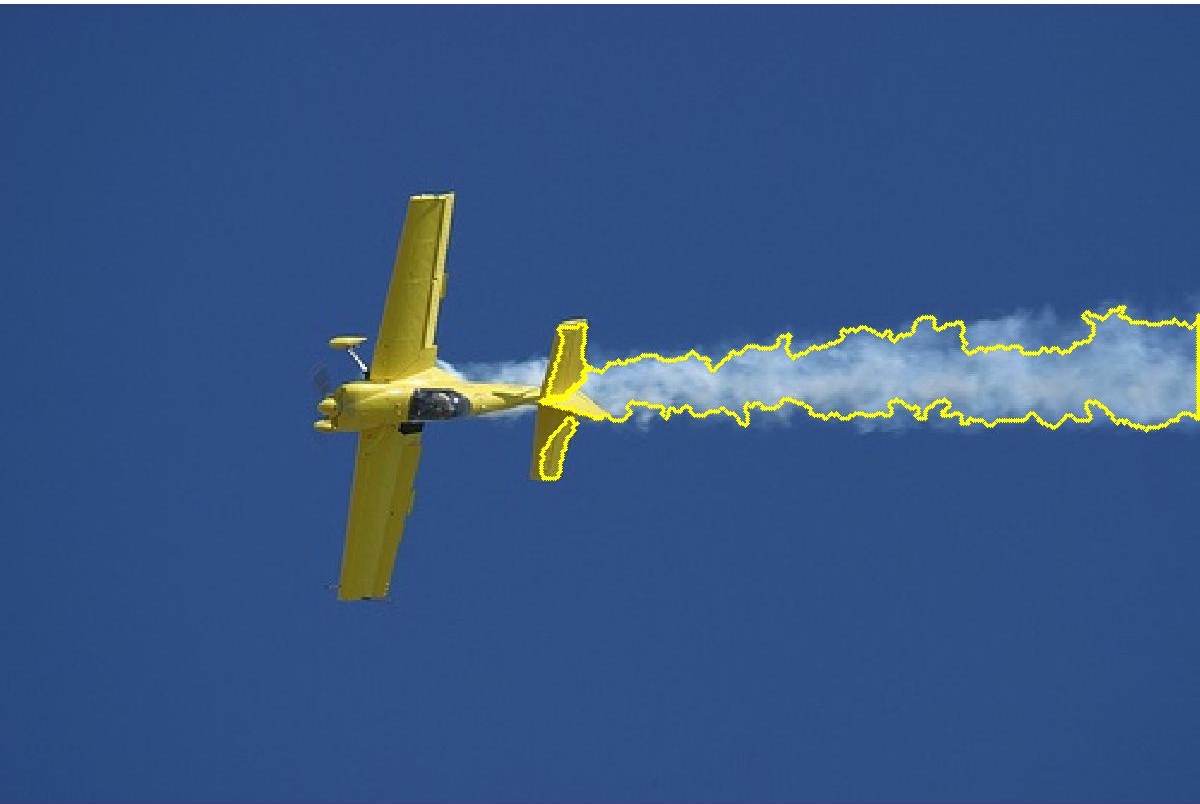}\\
\end{tabular}
\caption{
Example true positives from the ``aeroplane" class (\ie these
images contain at least one instance of aeroplane). On each image, the three
regions that are most similar to the top three prototypes learned by MIS-Boost are shown with
yellow boundaries (Best viewed in color). }
\label{fig:pascal_aeroplane_correct_classification}
\end{figure}

\begin{figure}
\centering
\begin{tabular}{ccc}
\includegraphics[scale=.10]{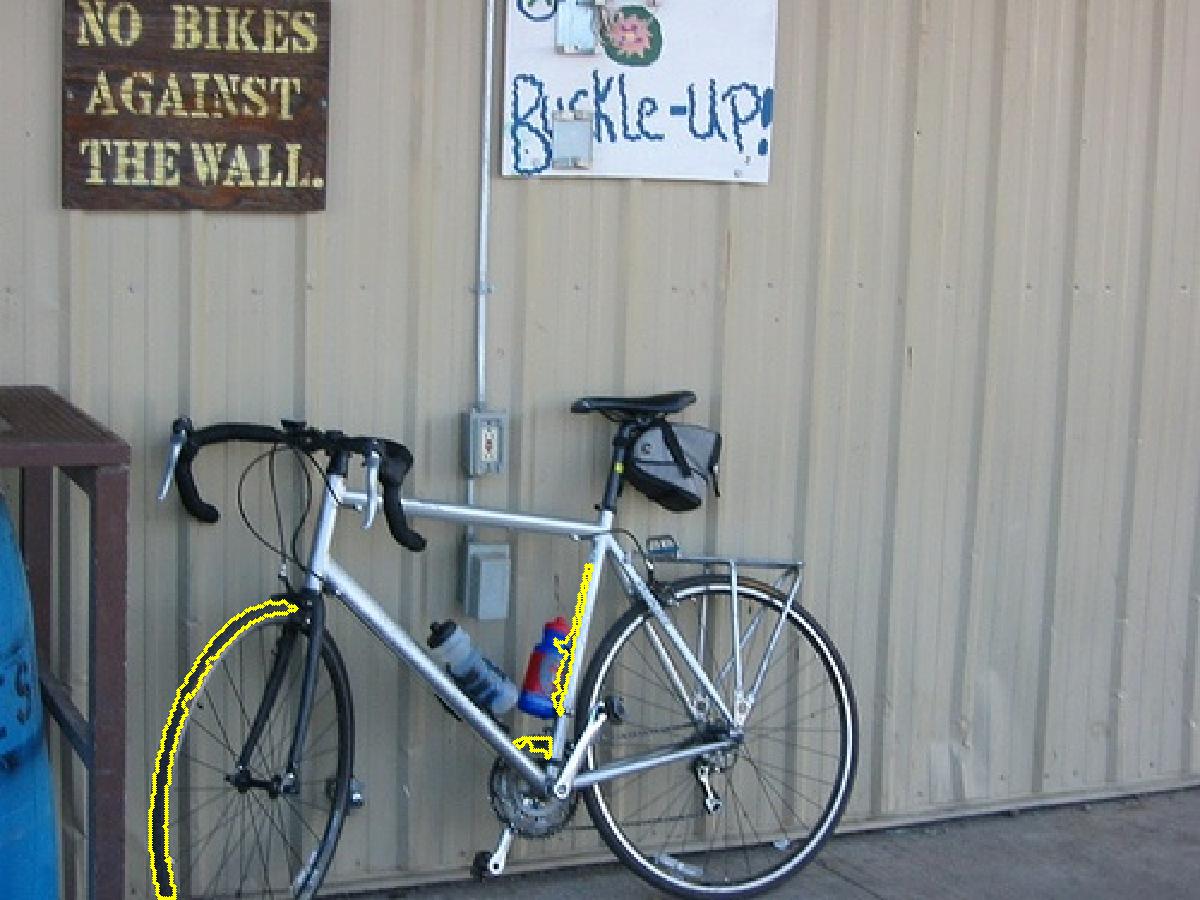}&
\includegraphics[scale=.10]{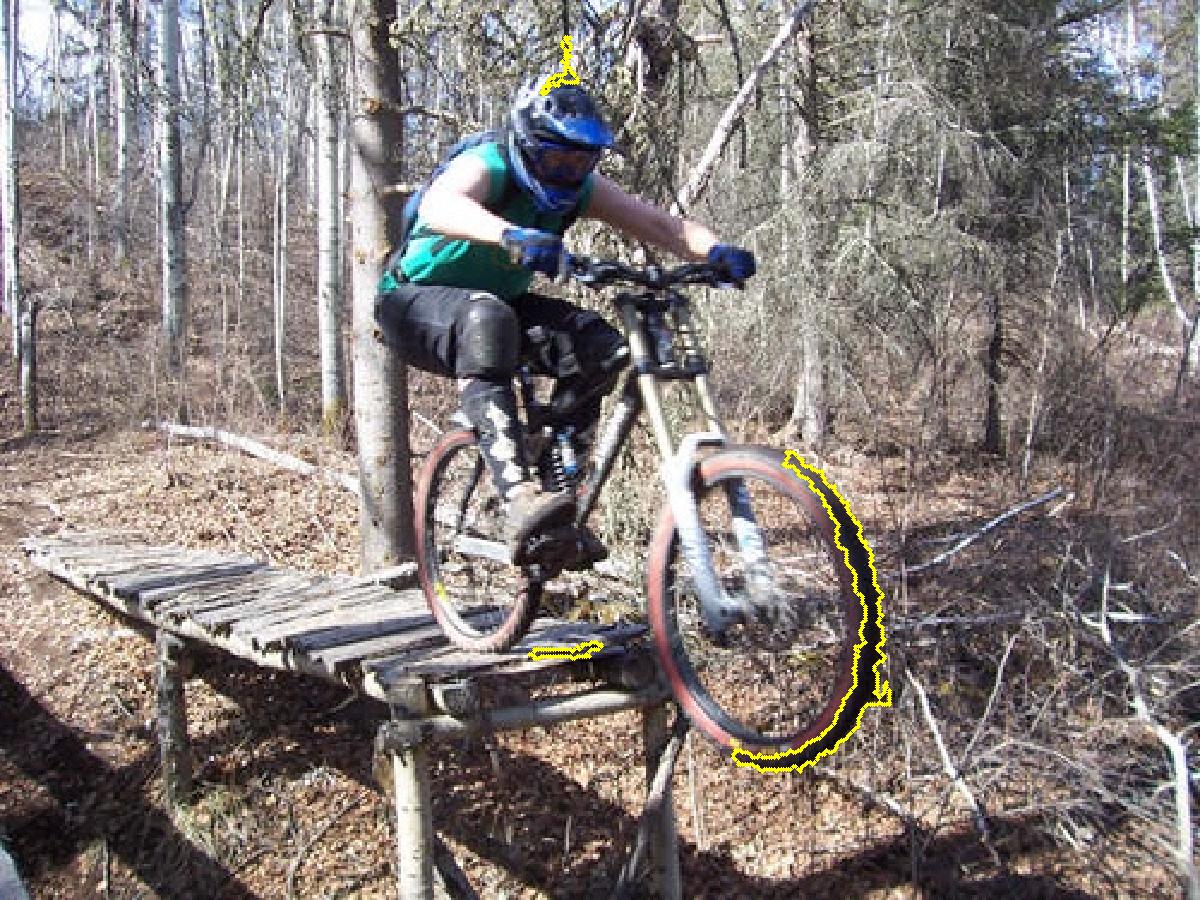}&
\includegraphics[scale=.10]{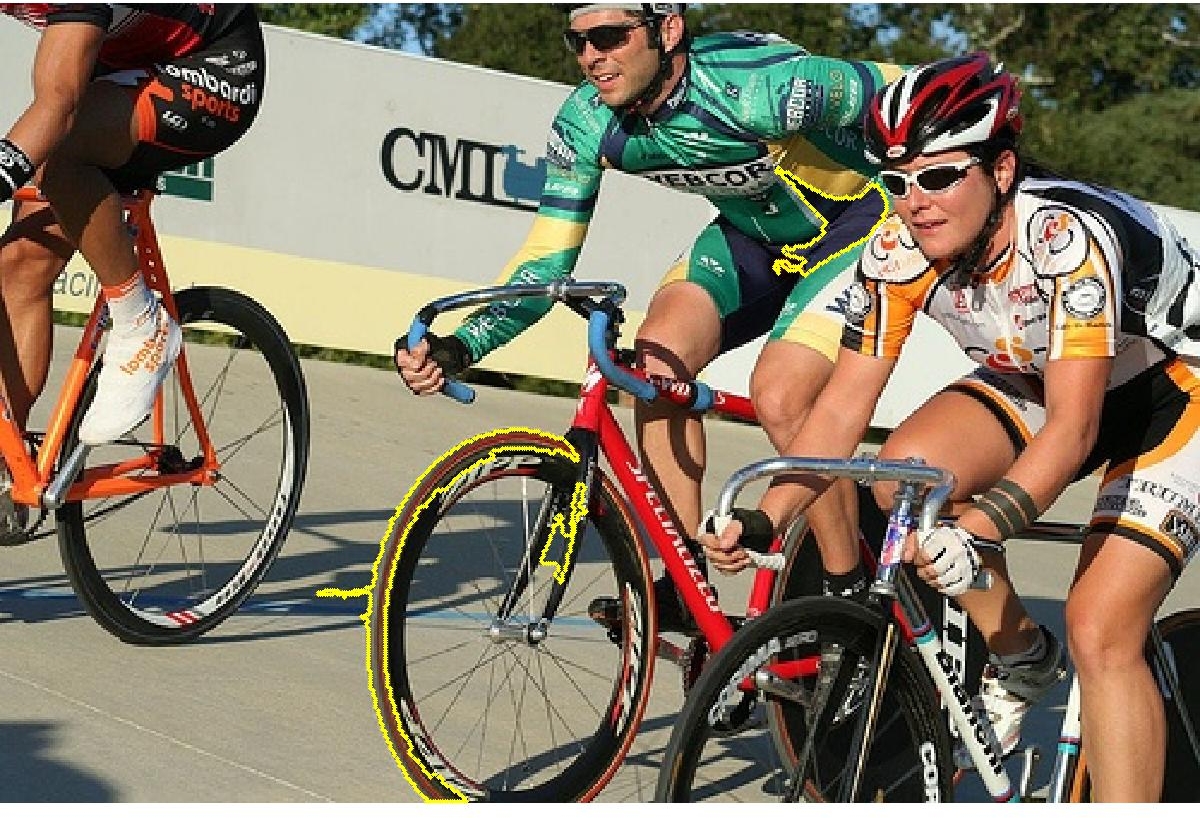}\\
\includegraphics[scale=.10]{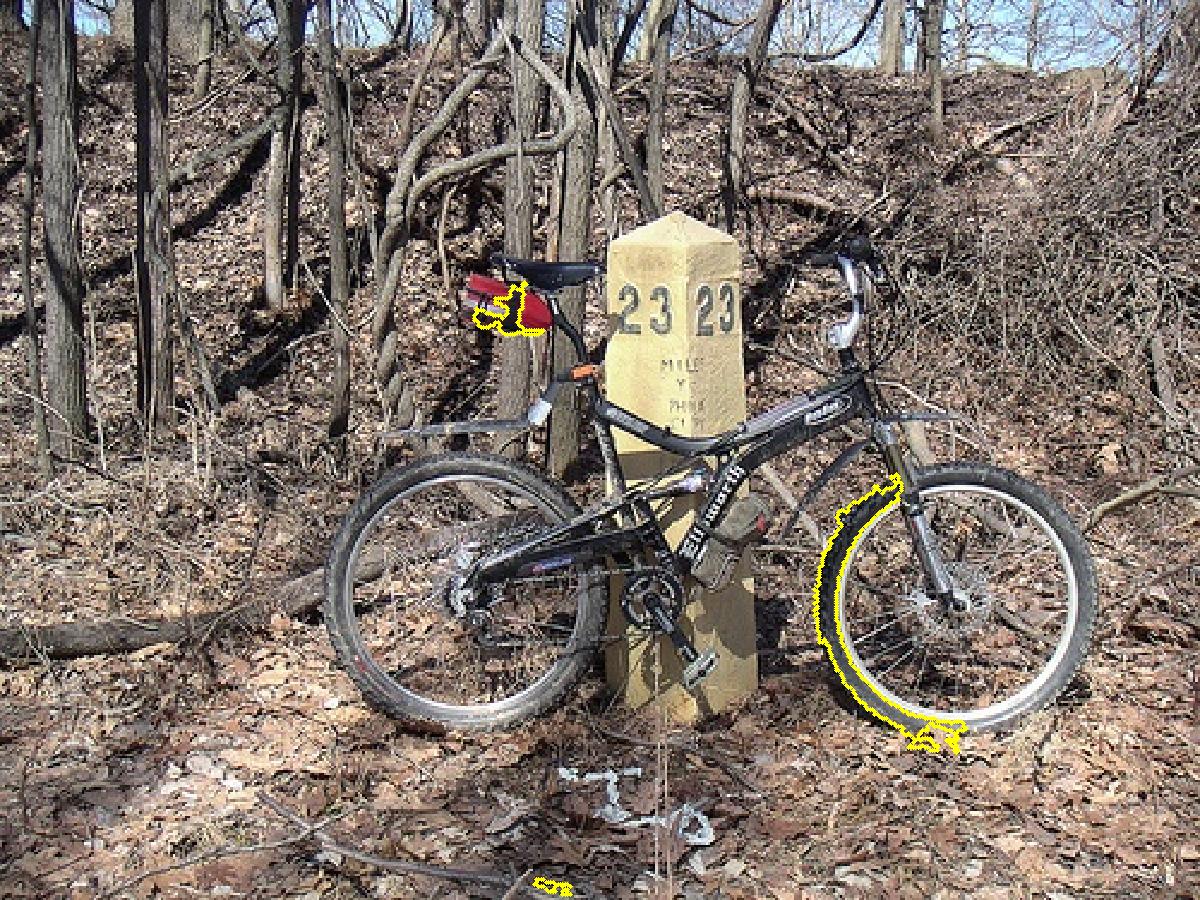}&
\includegraphics[scale=.10]{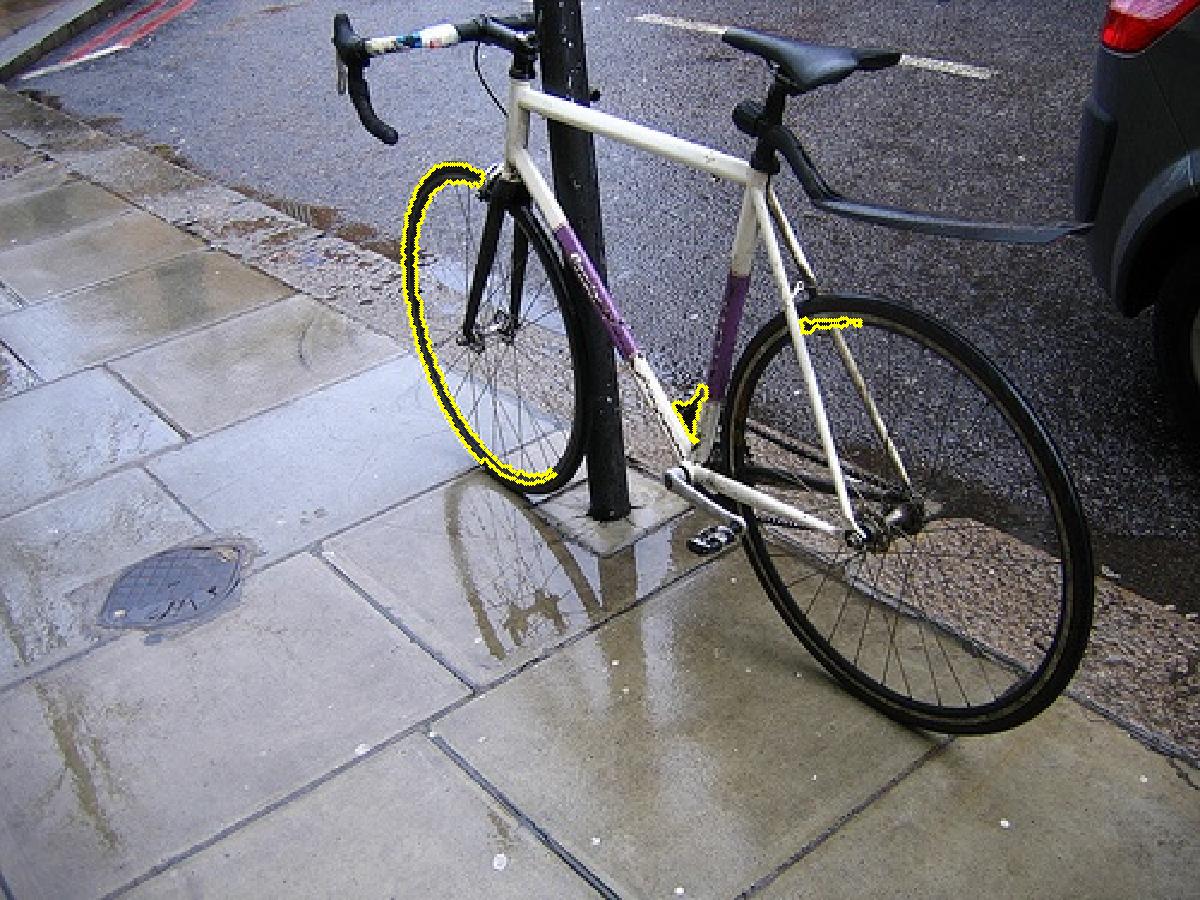}&
\includegraphics[scale=.10]{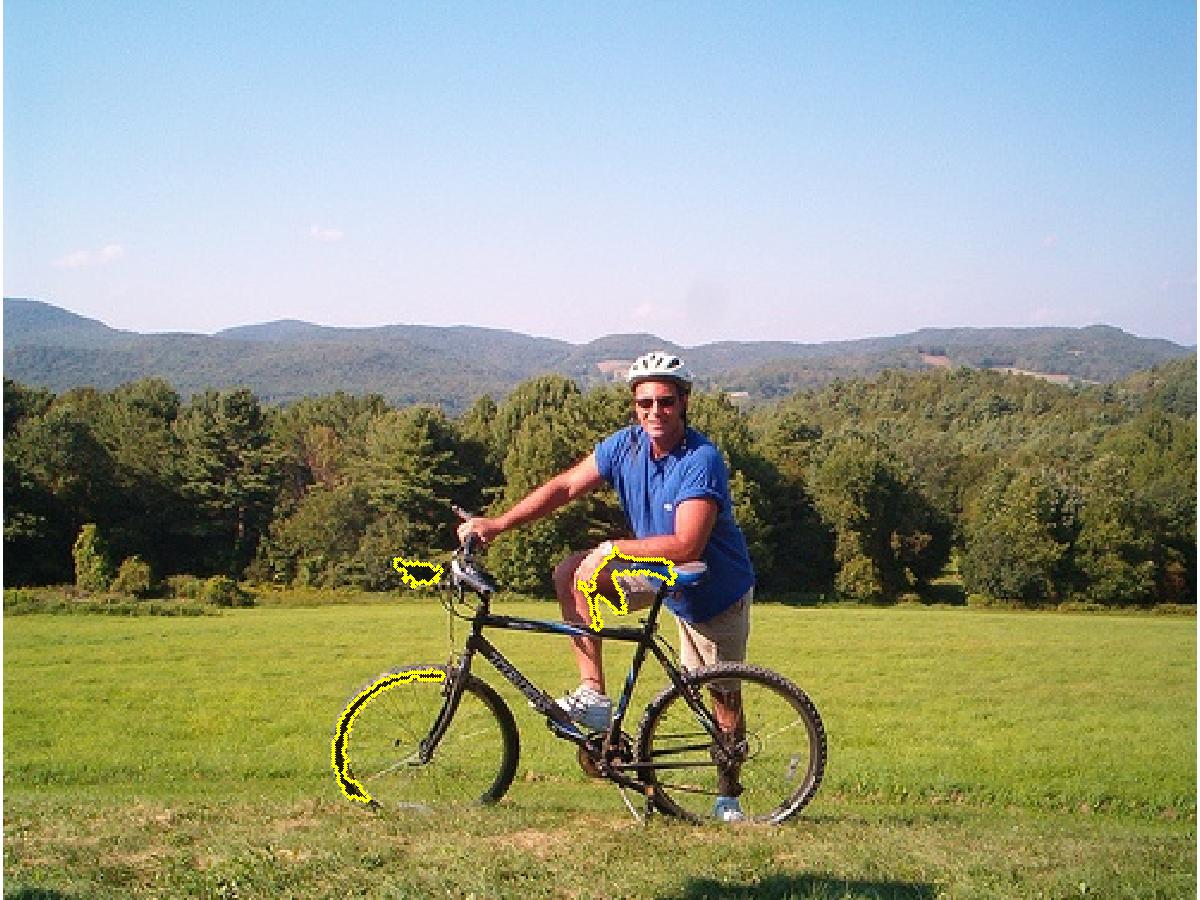}\\
\end{tabular}
\caption{
Example true positives from the ``bicycle" class. See caption of Figure
\ref{fig:pascal_aeroplane_correct_classification} for an explanation of the
yellow boundaries (Best viewed in color). }
\label{fig:pascal_bicycle_correct_classification}
\end{figure}

\begin{figure}
\centering
\begin{tabular}{ccc}
\includegraphics[scale=.10]{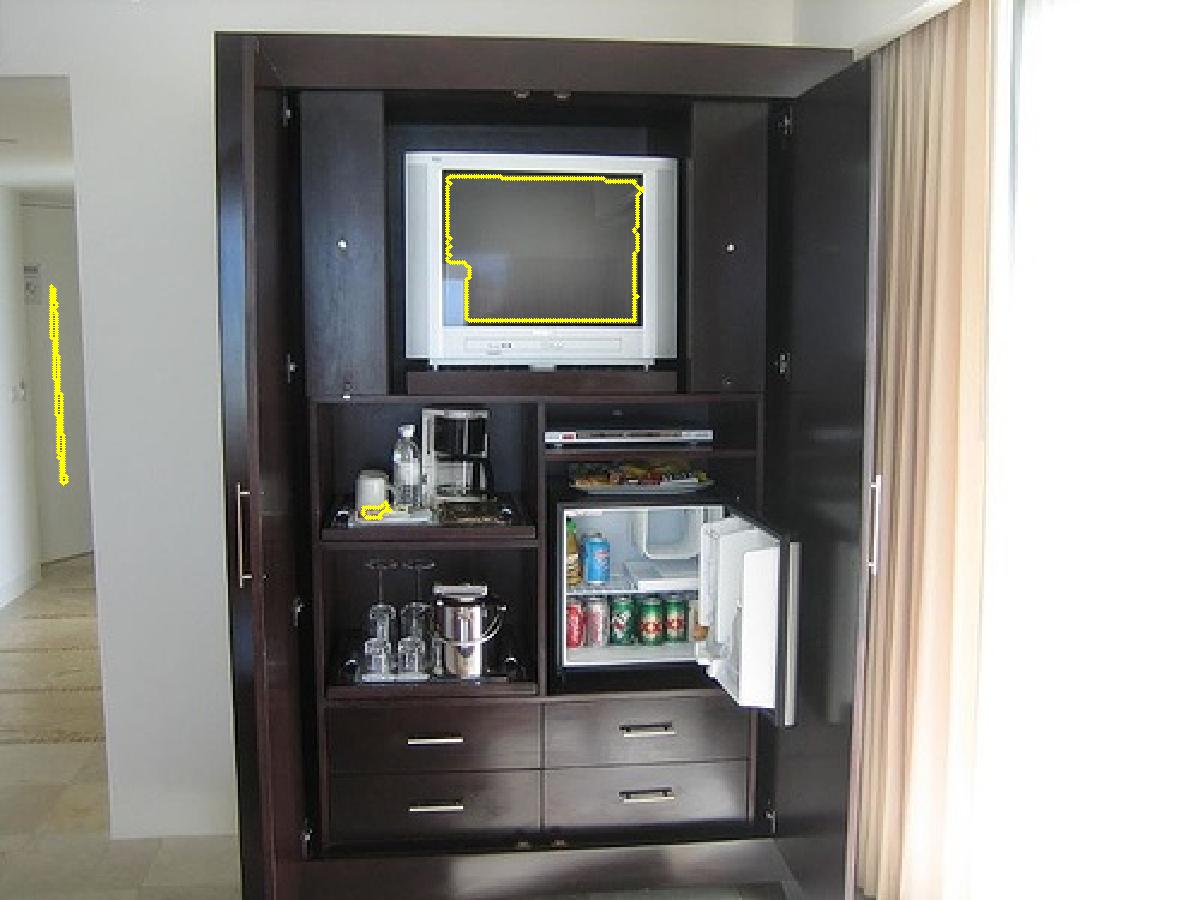}&
\includegraphics[scale=.10]{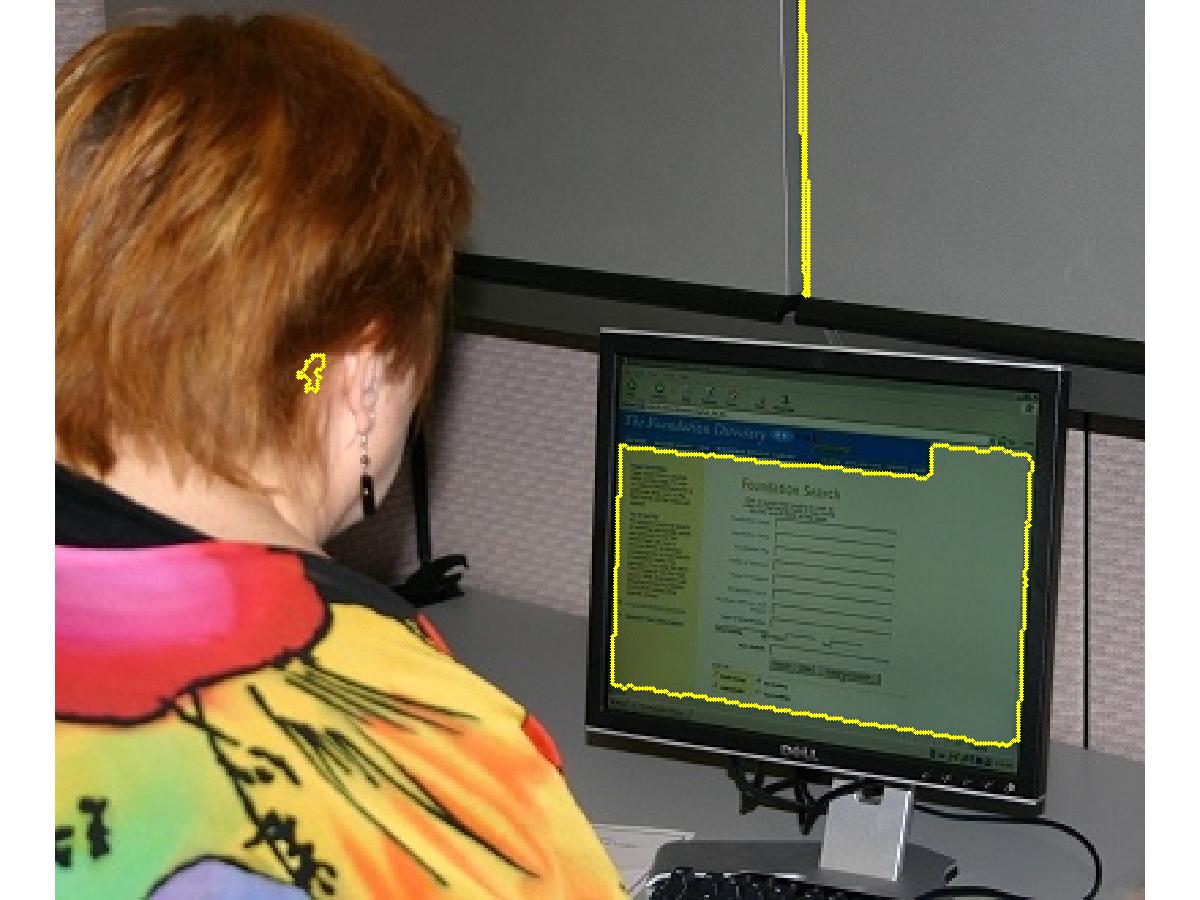}&
\includegraphics[scale=.10]{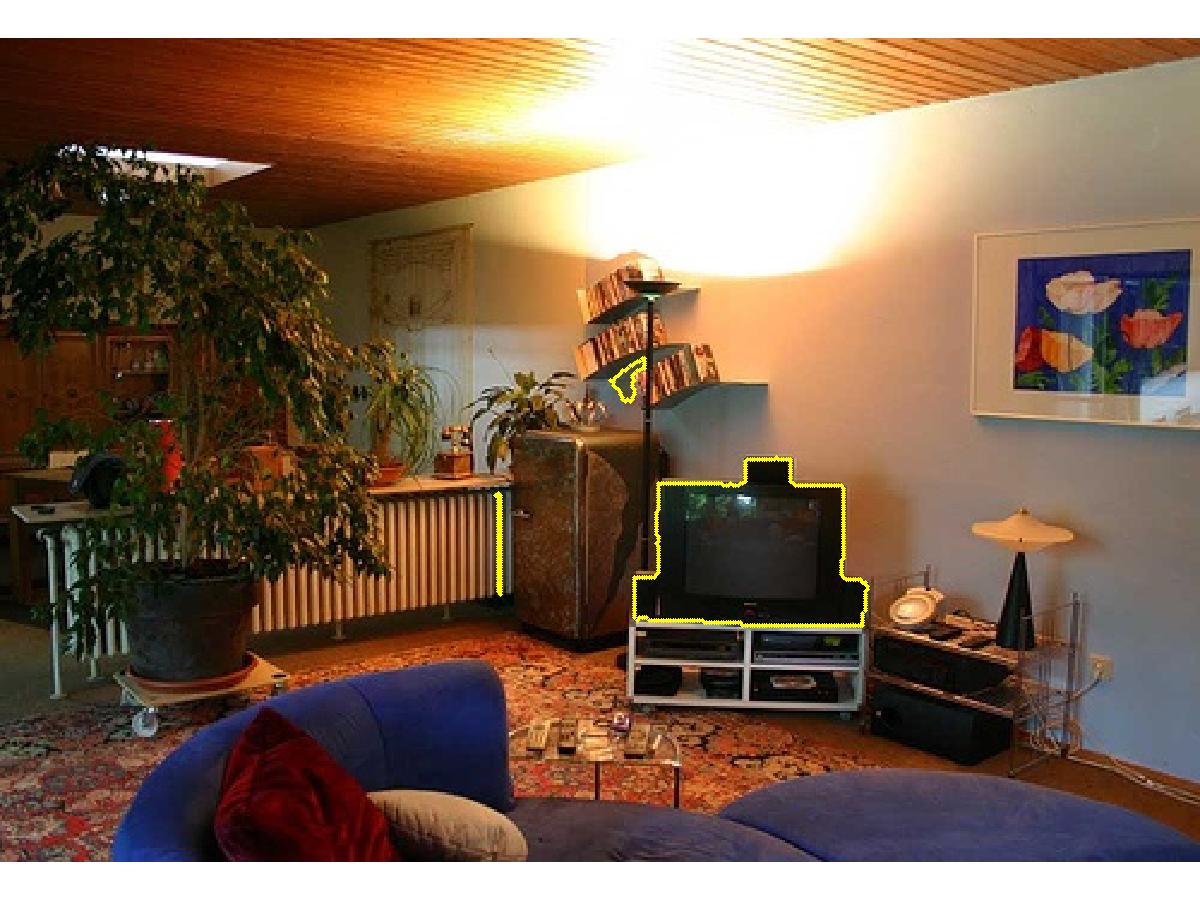}\\
\includegraphics[scale=.10]{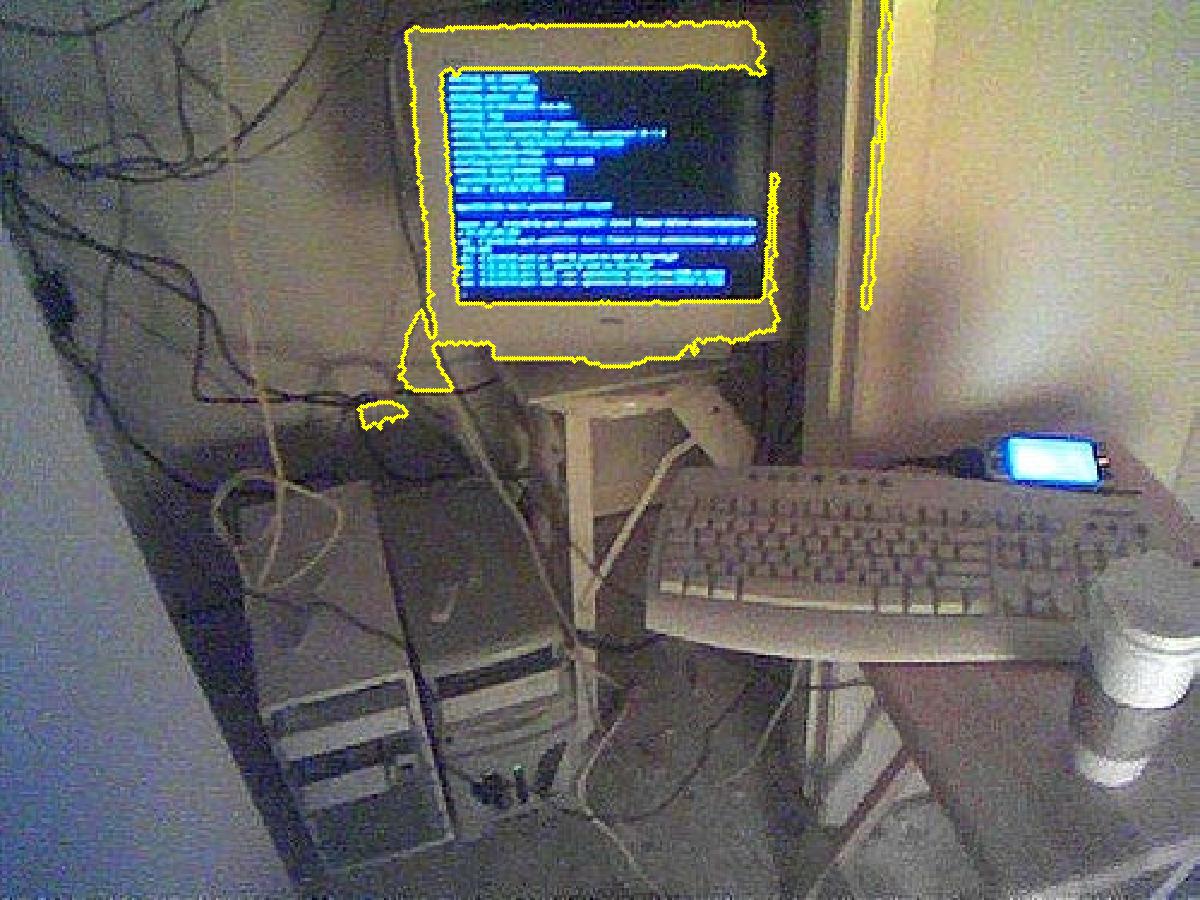}&
\includegraphics[scale=.10]{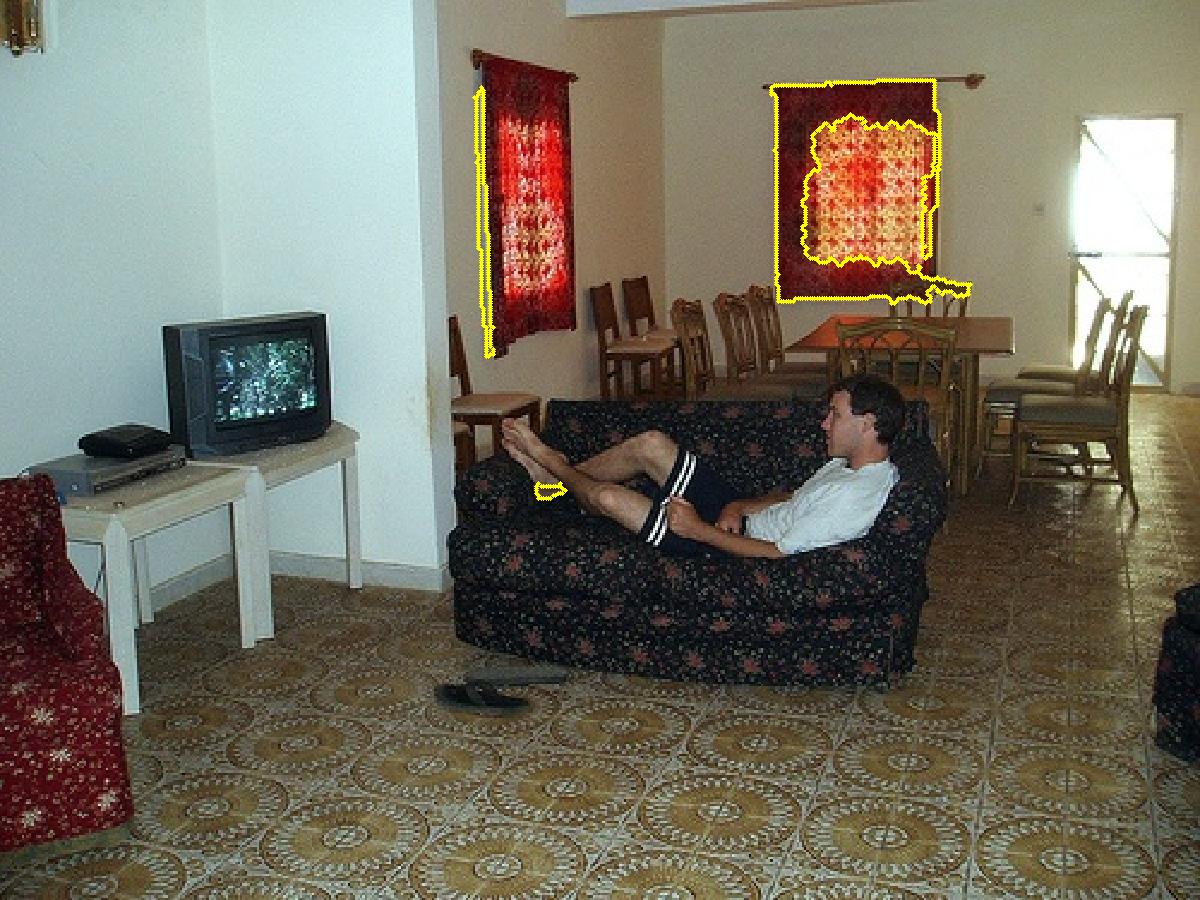}&
\includegraphics[scale=.10]{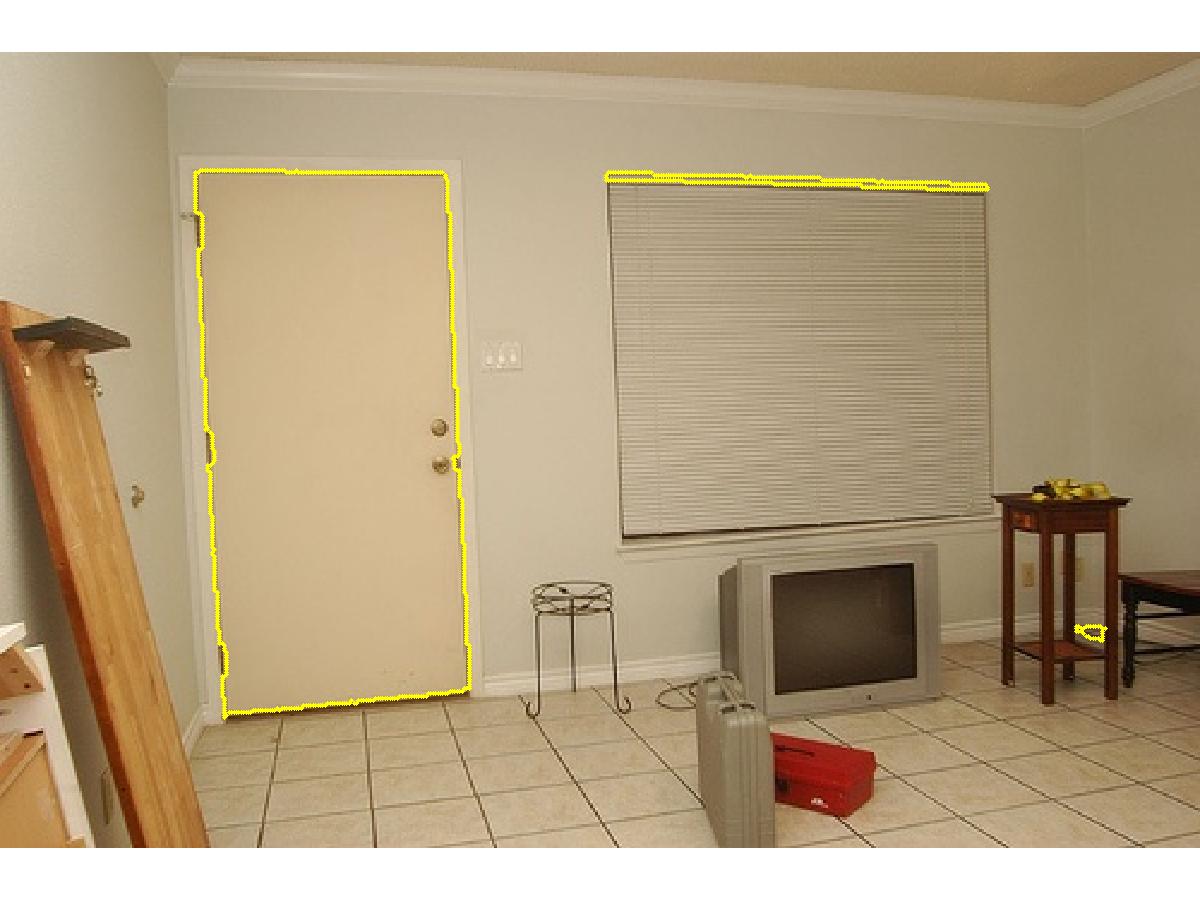}\\
\end{tabular}
\caption{
Example true positives from the ``tvmonitor" class. See caption of Figure
\ref{fig:pascal_aeroplane_correct_classification} for an explanation of the
yellow boundaries (Best viewed in color). }
\label{fig:pascal_tvmonitor_correct_classification}
\end{figure}

Next, we visualize the selected instances in each image that are closest to the learned prototypes. Figure \ref{fig:pascal_aeroplane_correct_classification} gives examples of true
positives from the aeroplane class. On each image, we show the top three instances,
i.e. regions, that are closest to the top $3$ prototypes learned by MIS-Boost.
These regions make the highest contribution to the correct classification of
their images. Similarly, Figure \ref{fig:pascal_bicycle_correct_classification} and
Figure \ref{fig:pascal_tvmonitor_correct_classification} present the same for the
``bicycle" and ``tvmonitor" classes, respectively.  As one can observe from the
images, the most discriminative regions usually overlap with the object of interest.
Occasionally, some wrong instances are selected as shown in the lower-middle,
and lower-right images of Figure
\ref{fig:pascal_tvmonitor_correct_classification}. Apparently, the ``tvmonitor"
classifier learned square-shaped or frame-shaped prototypes, and the
window in the lower-middle image, and the door in the lower-right image are good
selections for this prototype. We will make these challenging PASCAL datasets publicly available (online), in a MIL format, along
with the code needed to visualize selected instances. We hope that these
larger and more challenging datasets are used to compare MIL
methods in the future.

\section{Conclusion} \label{sec:conclusion}
We presented a new multiple instance learning (MIL) method that learns
discriminative instance prototypes by explicit instance selection in a
boosting framework. We argued that the following three design choices and/or
assumptions restrict the capacity of a MIL method: {\bf (i)} treating  the
prototype learning/choosing step and learning the final bag
classifier independently, {\bf (ii)}  restricting the prototypes to a discrete
set of instances from the training set, and {\bf (iii)} restricting the number
of selected-instances per bag. Our method, MIS-Boost, overcomes all three
restrictions by learning prototype-based base classifiers that are linearly
boosted. At each iteration of MIS-Boost, a prototype is learned so that it
maximally discriminates between the positive and negative bags, which are
themselves weighted according to how well they were discriminated in earlier
iterations. The number of total prototypes and the number of selected-instances
per bag are determined in a completely data-driven way. We showed
that our method outperforms state-of-the-art MIL methods on a number of benchmark
datasets. We also applied MIS-Boost to large-scale image classification,
where we showed that the automatically selected prototypes map to visually meaningful image regions.

%

\bibliographystyle{IEEEtran}
\bibliography{references,mendeley_MIL}

\begin{thebibliography}{10}
\providecommand{\url}[1]{#1}
\csname url@samestyle\endcsname
\providecommand{\newblock}{\relax}
\providecommand{\bibinfo}[2]{#2}
\providecommand{\BIBentrySTDinterwordspacing}{\spaceskip=0pt\relax}
\providecommand{\BIBentryALTinterwordstretchfactor}{4}
\providecommand{\BIBentryALTinterwordspacing}{\spaceskip=\fontdimen2\font plus
\BIBentryALTinterwordstretchfactor\fontdimen3\font minus
  \fontdimen4\font\relax}
\providecommand{\BIBforeignlanguage}[2]{{%
\expandafter\ifx\csname l@#1\endcsname\relax
\typeout{** WARNING: IEEEtran.bst: No hyphenation pattern has been}%
\typeout{** loaded for the language `#1'. Using the pattern for}%
\typeout{** the default language instead.}%
\else
\language=\csname l@#1\endcsname
\fi
#2}}
\providecommand{\BIBdecl}{\relax}
\BIBdecl

\bibitem{Andrews2003}
S.~Andrews, I.~Tsochantaridis, and T.~Hofmann, ``{Support vector machines for
  multiple-instance learning},'' \emph{NIPS}, pp. 561--568, 2003.

\bibitem{Dietterich1997}
T.~G. Dietterich, R.~H. Lathrop, and T.~Lozano-perez, ``{Solving the
  Multiple-Instance Problem with Axis-Parallel Rectangles},'' \emph{Artificial
  Intelligence}, vol.~89, pp. 31--71, 1997.

\bibitem{Zhang2002}
Q.~Zhang and S.~Goldman, ``{EM-DD: An improved multiple-instance learning
  technique},'' \emph{NIPS}, vol.~2, pp. 1073--1080, 2002.

\bibitem{Cholleti2006}
S.~Cholleti, S.~Goldman, and R.~Rahmani, ``{MI-Winnow: A New Multiple-Instance
  Learning Algorithm},'' \emph{International Conference on Tools with
  Artificial Intelligence}, pp. 336--346, 2006.

\bibitem{Vijayanarasimhan2008}
S.~Vijayanarasimhan and K.~Grauman, ``{Keywords to visual categories:
  Multiple-instance learning forweakly supervised object categorization},''
  \emph{CVPR}, pp. 1--8, 2008.

\bibitem{Fu2010a}
Z.~Fu, A.~Robles-Kelly, and J.~Zhou, ``{MILIS: Multiple Instance Learning with
  Instance Selection},'' \emph{{IEEE} TPAMI}, vol.~33, no.~10, pp. 1--20, 2010.

\bibitem{Chen2006}
Y.~Chen, J.~Bi, and J.~Z. Wang, ``{MILES: multiple-instance learning via
  embedded instance selection.}'' \emph{{IEEE} TPAMI}, vol.~28, no.~12, pp.
  1931--47, 2006.

\bibitem{Foulds2008}
J.~Foulds and E.~Frank, ``{Revisiting multi-instance learning via embedded
  instance selection},'' \emph{Lecture Notes in Computer Science}, 2008.

\bibitem{Stikic2009}
M.~Stikic and B.~Schiele, ``{Activity recognition from sparsely labeled data
  using multi-instance learning},'' \emph{Location and Context Awareness}, pp.
  156--173, 2009.

\bibitem{Babenko2010}
B.~Babenko, M.-H. Yang, and S.~Belongie, ``{Visual Tracking with Online
  Multiple Instance Learning},'' \emph{{IEEE} TPAMI}, pp. 983--990, 2010.

\bibitem{Leistner2010}
C.~Leistner, A.~Saffari, and H.~Bischof, ``{MIForests: Multiple-Instance
  Learning with Randomized Trees},'' \emph{ECCV}, pp. 29--42, 2010.

\bibitem{milboost:nips2007}
P.~Viola, J.~C. Platt, and C.~Zhang, ``{Multiple Instance Boosting for Object
  Detection},'' in \emph{NIPS}, 2007.

\bibitem{Zhang2008}
C.~Zhang and P.~Viola, ``{Multiple-instance pruning for learning efficient
  cascade detectors},'' in \emph{NIPS}, 2008.

\bibitem{Vezhnevets2010}
A.~Vezhnevets and J.~Buhmann, ``{Towards weakly supervised semantic
  segmentation by means of multiple instance and multitask learning},'' in
  \emph{CVPR}, 2010, pp. 3249--3256.

\bibitem{Ray2005}
S.~Ray and M.~Craven, ``{Supervised versus multiple instance learning: An
  empirical comparison},'' in \emph{International Conference on Machine
  Learning}, 2005, pp. 697--704.

\bibitem{Keeler1990}
J.~Keeler, D.~Rumelhart, and W.~Leow, ``{Integrated segmentation and
  recognition of hand-printed numerals},'' in \emph{NIPS}, 1990, pp. 557--563.

\bibitem{Cheung2006}
P.-M. Cheung and J.~T. Kwok, ``{A regularization framework for
  multiple-instance learning},'' \emph{International Conference on Machine
  Learning}, no.~1, pp. 193--200, 2006.

\bibitem{Zhou2007a}
Z.-H. Zhou and J.-M. Xu, ``{On the relation between multi-instance learning and
  semi-supervised learning},'' \emph{ICML}, no. 1997, pp. 1167--1174, 2007.

\bibitem{Maron1998}
O.~Maron and T.~Lozano-P\'{e}rez, ``{A framework for multiple-instance
  learning},'' in \emph{NIPS}, 1998, pp. 570--576.

\bibitem{Rahmani2008}
R.~Rahmani, S.~a. Goldman, H.~Zhang, S.~R. Cholleti, and J.~E. Fritts,
  ``{Localized content-based image retrieval.}'' \emph{{IEEE} TPAMI}, vol.~30,
  no.~11, pp. 1902--12, 2008.

\bibitem{Wang2000}
J.~Wang and J.~Zucker, ``{Solving the multiple-instance problem: A lazy
  learning approach},'' in \emph{ICML}, no. 1994, 2000, pp. 1119--1126.

\bibitem{Gartner2002}
T.~Gartner, P.~Flach, A.~Kowalczyk, and A.~Smola, ``{Multi-instance kernels},''
  in \emph{ICML}, 2002, pp. 179--186.

\bibitem{Chen2004}
Y.~Chen and J.~Wang, ``{Image categorization by learning and reasoning with
  regions},'' \emph{JMLR}, vol.~5, pp. 913--939, 2004.

\bibitem{milboost}
P.~Viola, J.~C. Platt, and C.~Zhang, ``{Multiple Instance Boosting for Object
  Detection},'' in \emph{NIPS}, 2007.

\bibitem{Fu2008}
Z.~Fu, ``{Fast multiple instance learning via l1, 2 logistic regression},''
  \emph{ICPR}, pp. 1--4, 2008.

\bibitem{Fu2009}
Z.~Fu and A.~Robles-Kelly, ``{An instance selection approach to Multiple
  Instance Learning},'' in \emph{CVPR}.\hskip 1em plus 0.5em minus 0.4em\relax
  IEEE Computer Society, 2009, pp. 911--918.

\bibitem{statistical_view_of_boosting}
J.~Friedman, T.~Hastie, and R.~Tibshirani, ``{Additive Logistic Regression: A
  Statistical View of Boosting},'' \emph{The Annals of Statistics}, vol.~28,
  no.~2, pp. 337--407, 2000.

\bibitem{Zhou2009}
Z.-H. Zhou, Y.-Y. Sun, and Y.-F. Li, ``{Multi-instance learning by treating
  instances as non-I.I.D. samples},'' \emph{ICML}, pp. 1--8, 2009.

\bibitem{milis:pami2010}
Z.~Fu, A.~Robles-Kelly, and J.~Zhou, ``Milis: Multiple instance learning with
  instance selection,'' \emph{{IEEE} TPAMI}, pp. 958 -- 977, Aug 2010.

\bibitem{GehlerChapelle_AISTATS2007}
P.~Gehler and O.~Chapelle, ``Deterministic annealing for multiple-instance
  learning,'' in \emph{Proc. of the 11th Int. Conference on Artificial
  Intelligence and Statistics}, Meila, M., and X.~Shen, Eds., March 2007, pp.
  123--130.

\bibitem{sift}
D.~G. Lowe, ``Distinctive image features from scale-invariant keypoints,''
  \emph{IJCV}, no.~2, pp. 91 -- 110, 2004.

\bibitem{akbas:accv2009}
E.~Akbas and N.~Ahuja, ``{From Ramp Discontinuities to Segmentation Tree},'' in
  \emph{Asian Conference on Computer Vision (ACCV'09)}.\hskip 1em plus 0.5em
  minus 0.4em\relax Xi'an, China: Springer, 2009, pp. 123 --134.

\bibitem{Perronnin2010}
F.~Perronnin, J.~S\'{a}nchez, and T.~Mensink, ``{Improving the Fisher Kernel
  for Large-Scale Image Classification},'' in \emph{ECCV}, vol. 6314.\hskip 1em
  plus 0.5em minus 0.4em\relax Springer Berlin / Heidelberg, 2010, pp.
  143--156--156.

\end{thebibliography}

\end{document}